\RequirePackage{iftex}
\ifPDFTeX
\pdfoutput=1
\fi

\documentclass[11pt]{article}

\usepackage[final]{acl}

\usepackage{times}
\usepackage{latexsym}

\ifPDFTeX
\usepackage[T1]{fontenc}

\usepackage[utf8]{inputenc}
\usepackage{CJKutf8}
\newcommand{\cn}[1]{{\begin{CJK*}{UTF8}{gbsn}#1\end{CJK*}}}

\else
\usepackage[no-math]{fontspec}
\usepackage{xparse}
\defaultfontfeatures{Ligatures=TeX}
\newfontfamily\cjkfont{FandolSong-Regular.otf}[Script=CJK]
\newcommand{\cn}[1]{{\begingroup\cjkfont\XeTeXlinebreaklocale "zh"\XeTeXlinebreakskip = 0pt plus 0.1em\relax #1\endgroup}}
\NewDocumentEnvironment{cjk}{}{\begingroup\cjkfont\XeTeXlinebreaklocale "zh"\XeTeXlinebreakskip = 0pt plus 0.1em\relax}{\endgroup}
\NewDocumentEnvironment{CJK*}{m m}{\begingroup\cjkfont\XeTeXlinebreaklocale "zh"\XeTeXlinebreakskip = 0pt plus 0.1em\relax}{\endgroup}
\fi

\usepackage{microtype}

\ifPDFTeX
\IfFileExists{inconsolata.sty}{\usepackage{inconsolata}}{}
\fi

\usepackage{graphicx}

\usepackage{url}
\IfFileExists{xspace.sty}{\usepackage{xspace}}{\newcommand{\xspace}{}}
\IfFileExists{subfigure.sty}{\usepackage{subfigure}}{}
\IfFileExists{multirow.sty}{\usepackage{multirow}}{\newcommand{\multirow}[3]{##3}}
\IfFileExists{enumitem.sty}{\usepackage{enumitem}}{}
\IfFileExists{stmaryrd.sty}{\usepackage{stmaryrd}}{}
\usepackage{array}
\IfFileExists{threeparttable.sty}{\usepackage{threeparttable}}{}
\IfFileExists{blindtext.sty}{\usepackage{blindtext}}{}
\usepackage{amssymb}

\IfFileExists{algorithm2e.sty}{\usepackage[ruled,vlined,linesnumbered,longend]{algorithm2e}}{}
\IfFileExists{placeins.sty}{\usepackage{placeins}}{}
\IfFileExists{tikz.sty}{\usepackage{tikz}}{}

\IfFileExists{soul.sty}{\usepackage{soul}}{\newcommand{\sethlcolor}[1]{}}
\usepackage{graphicx}
\usepackage{amsmath}
\usepackage{booktabs}
\IfFileExists{lipsum.sty}{\usepackage{lipsum}}{}
\usepackage{listings}
\usepackage{xcolor}
\sethlcolor{blue!10}
\usepackage{colortbl}
\IfFileExists{bbm.sty}{\usepackage{bbm}}{}

\IfFileExists{epigraph.sty}{
  \usepackage{epigraph}
  
  \setlength \epigraphrule {0pt}
  \setlength\epigraphwidth{.34\textwidth}
}{}

\IfFileExists{cleveref.sty}{
  \usepackage{cleveref}
  \crefformat{section}{\S##2##1##3} 
  \crefformat{subsection}{\S##2##1##3}
  \crefformat{subsubsection}{\S##2##1##3}
}{}

\usepackage{hyperref}

\newcommand{\method}{{\normalfont\textsc{HistLens}}\xspace}

\newcommand{\githubicon}{%
  \raisebox{-0.18em}{%
    \tikz[scale=0.14,baseline={(0,-0.2)}]{
      \fill[black] (0,0) circle (1.15);
      \fill[black] (-0.58,0.78) -- (-0.16,1.72) -- (0.16,0.82) -- cycle;
      \fill[black] (0.58,0.78) -- (0.16,1.72) -- (-0.16,0.82) -- cycle;
    }%
  }%
}

\newcommand{\translationline}[1]{\noindent{\small\itshape #1}\par}
\newcommand{\validationtranslationline}[1]{\noindent{\small\textit{Translation.} \itshape #1}\par}
\newcommand{\analysisline}[1]{\noindent{\small #1}\par}

\newcommand{\validationexample}[5]{%
  \item \textbf{#1}\quad \texttt{act}=#2\par
  \noindent\cn{#3}\par
  \validationtranslationline{#4}%
  \analysisline{\textit{Analysis.} #5}%
}
\newcommand{\indexedquote}[3]{%
  \item \textbf{#1}\par
  \noindent\cn{#2}\par
  \translationline{#3}%
}
\newcommand{\evidencegroup}[1]{\smallskip\noindent\textit{#1}\par}
\newenvironment{evidencelist}{%
  \begin{enumerate}
  \setlength{\itemsep}{0.35em}
  \setlength{\topsep}{0.2em}
  \setlength{\partopsep}{0pt}
  \setlength{\parskip}{0pt}
  \setlength{\parsep}{0pt}
}{%
  \end{enumerate}
}

\definecolor{sgreen}{HTML}{F3FADF}  
\definecolor{mgreen}{HTML}{E0EAB5}  
\definecolor{dgreen}{HTML}{CDDC8C}  
\definecolor{ddgreen}{HTML}{B8CF61} 

\usepackage{amsmath,amsfonts,bm}

\def\eqref#1{equation~\ref{#1}}

\def\1{\bm{1}}

\DeclareMathAlphabet{\mathsfit}{\encodingdefault}{\sfdefault}{m}{sl}
\SetMathAlphabet{\mathsfit}{bold}{\encodingdefault}{\sfdefault}{bx}{n}

\title{
HistLens: Mapping Idea Change across
\\ Concepts and Corpora
}

\author{
  Yi Jing$^{\spadesuit\diamondsuit}$\thanks{Equal contribution.}
  \quad Weiyun Qiu$^{\heartsuit}$\footnotemark[1]
  \quad Yihang Peng$^{\clubsuit}$
  \quad Zhifang Sui$^{\diamondsuit}$\thanks{Corresponding author.} \\
  $^\spadesuit$Department of Computer Science and Technology, Tsinghua University, China \\
  $^\heartsuit$School of History, Nanjing University, China \\
  $^\clubsuit$Department of Chinese Language and Literature, Tsinghua University, China \\
  $^\diamondsuit$State Key Laboratory of Multimedia Information Processing, \\
  School of Computer Science, Peking University, China \\
  \texttt{jingy22@mails.tsinghua.edu.cn}, \texttt{szf@pku.edu.cn}
}

\begin{document}
\maketitle

\begin{abstract}
Language change both reflects and shapes social processes, and the semantic evolution of foundational concepts provides a measurable trace of historical and social transformation. Despite recent advances in diachronic semantics and discourse analysis, existing computational approaches often (i) concentrate on a single concept or a single corpus, making findings difficult to compare across heterogeneous sources, and (ii) remain confined to surface lexical evidence, offering insufficient computational and interpretive granularity when concepts are expressed implicitly.
We propose \method, a unified, SAE-based framework for multi-concept, multi-corpus conceptual-history analysis. The framework decomposes concept representations into interpretable features and tracks their activation dynamics over time and across sources, yielding comparable conceptual trajectories within a shared coordinate system. Experiments on long-span press corpora show that \method supports cross-concept, cross-corpus computation of patterns of idea evolution and enables implicit concept computation. By bridging conceptual modeling with interpretive needs, \method broadens the analytical perspectives and methodological repertoire available to social science and the humanities for diachronic text analysis.
\end{abstract}

\begin{center}
\vspace{0.25em}
\setlength{\tabcolsep}{0.45em}
\begin{tabular}{@{}rl@{}}
\githubicon\enspace \textbf{Code} & \href{https://github.com/LeoJ-xy/HistLens}{\textcolor{blue}{LeoJ-xy/HistLens}}
\end{tabular}
\vspace{0.2em}
\end{center}

\section{Introduction}

\begin{flushright}
\begin{minipage}{0.95\linewidth}
\begin{quote}\itshape
``Concepts are both indicators and factors of historical change.''
\end{quote}
\vspace{-0.75em}
\hfill---\textsc{Reinhart Koselleck}
\end{minipage}
\end{flushright}

Language and society co-evolve as two intertwined trajectories. Language change not only reflects but also shapes social processes, and the semantic evolution of fundamental concepts constitutes an observable marker of historical transformation. Concepts are more than lexical items: they are configurations of meaning, association, and argumentative roles embedded in social contexts, continuously reworked as social structures and public discourse shift. For computational social science and digital humanities, an important goal is therefore to characterize conceptual change in large-scale diachronic corpora, and to measure it in ways that support historical interpretation and social theory---ultimately yielding new, valuable insights for both social science and historiography.

\begin{figure}[t]
    \centering
    \includegraphics[width=0.98\linewidth]{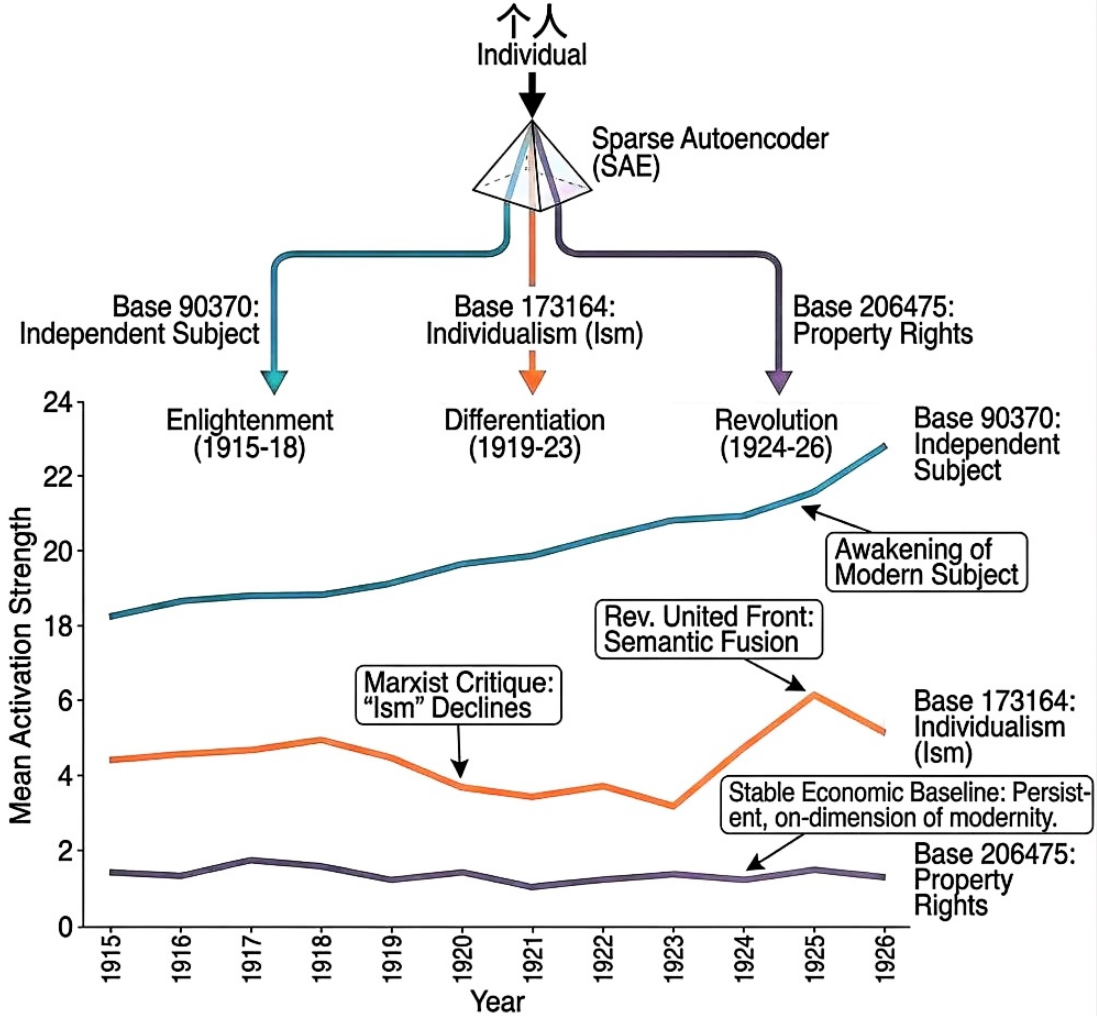}
    \caption{In \emph{New Youth}, \emph{individual} is not unitary but decomposes into sub-semantics such as \textit{independent subjecthood}, \textit{individualism as discourse}, and \textit{property rights/economic individuality}, each with a partly independent trajectory over 1915--1926. \method identifies such internal conceptual heterogeneity and diachronic reconfiguration.}
    \label{fig:cominter}
\end{figure}

Recent years have witnessed substantial progress in NLP for diachronic semantics and discourse analysis, including work on lexical semantic change\cite{Hamilton2016Diachronic, PeritiTahmasebi2024BeyondTwoPeriods}, topic evolution\cite{BleiLafferty2006DTM, Dieng2019DynamicETM, SirinLippincott2024DynamicETM, James2024DTMEvalACL}, and the study of stances and frames in public discourse\cite{Otmakhova2024FramingSurveyACL, Irani2025DALiSM}. Yet, integrating these advances into a scalable, comparable, and interpretable paradigm for studying conceptual semantic evolution remains challenging.

First, \textbf{scalability and comparability across concepts and across corpora are still limited}. A large portion of existing work focuses on a single concept, a small set of keywords, or a single corpus\cite{Tang2023SenseDistributionLSC, Gribomont2023SDK, Matthews2025KSparseLSC}. Even when methods scale technically\cite{James2024DTMEvalACL, Ma2025PeoplesDailyIdeology}, their outputs are often not directly comparable across different concepts or across heterogeneous sources. This makes it difficult to address canonical questions such as: Do multiple concepts co-evolve in coordinated ways? Which changes are shared across corpora, and which are products of specific contexts?

Second, \textbf{existing approaches often fall short in characterizing implicit concepts}. Many methods remain centered on keywords and surface co-occurrence patterns, making it difficult to capture fine-grained intellectual and social evolution that is not explicitly expressed in canonical terms\cite{TimkeyVanSchijndel2021RogueDimensions, James2024DTMEvalACL, Otmakhova2024FramingSurveyACL, Irani2025DALiSM}, thereby limiting analytical depth. For historians and social scientists, this yields two direct limitations: (i) ``concept change'' can be misread as mere lexical replacement or stylistic fluctuation, obscuring continuity and turns across shifting discursive strategies; and (ii) overlooking implicit expressions introduces bias in source selection and interpretation, so that identifying key turning points relies more on researchers' experience than on a systematic and objective evidential chain.

To address these challenges, we propose \method, a \textbf{unified conceptual-history analysis framework} for \textbf{multiple concepts} and \textbf{multiple corpora}. The framework is built on a sparse, feature-structured representational space: we decompose dynamic semantic representations into interpretable features, and recast conceptual inquiry as tracking the activation dynamics of these features across time and across sources. By anchoring different concepts within a shared sparse feature coordinate system, the framework enables consistent measurement and principled comparison across heterogeneous corpora. Empirically, we apply \method to long-span press corpora, demonstrating its scalability and its capacity to bridge computational modeling with humanistic interpretation. 

\method is the first framework to enable unified measurement and comparable analysis of multiple concepts across multiple corpora within a shared, interpretable sparse feature space. This moves the study of concept evolution beyond keyword-level fluctuations by decomposing change into quantifiable and explainable semantic components. Moreover, it allows us to compute implicit semantic and ideational trajectories in texts where a concept is not explicitly named, offering a more systematic and interpretable quantitative methodology, and also a new analytic lens for humanistic close reading and social-scientific inquiry.

\section{Related Works}

\subsection{Diachronic Semantic Change}
A major computational route to conceptual history operationalizes ``concept evolution'' as diachronic semantic change, from long-run cultural and lexical trend quantification in massive digitized corpora \citep{Michel2011Culturomics,Hamilton2016Diachronic} to contextual-representation pipelines that extract contextual word embeddings from masked language models and compare them across time via similarity or clustering \citep{Devlin2019BERT}. Beyond term-level comparisons, related work models change via time-specific sense distributions \citep{Tang2023SenseDistributionLSC}, cross-temporal context perturbations \citep{AidaBollegala2023SwapPredict}, and multi-period diachronic sense induction \citep{PeritiTahmasebi2024BeyondTwoPeriods}; in digital humanities, semantic-difference keyword methods target cross-corpus divergence and interpretable ``sites of semantic struggle'' \citep{Gribomont2023SDK}. Complementarily, concept/topic discovery examines how new semantic regions emerge and diffuse \citep{MaNyarko2025EmergingConcepts}, while dynamic topic models \citep{BleiLafferty2006DTM}, neural extensions such as the dynamic embedded topic model \citep{Dieng2019DynamicETM}, literary-historical operationalizations of that line of work \citep{SirinLippincott2024DynamicETM}, and recent evaluations \citep{James2024DTMEvalACL} support more reliable temporal topic comparisons.

\subsection{Diachronic Modeling of Idea and Discourse}
Beyond lexical meaning, computational social science and digital humanities study how stance, framing, and discourse regimes shift over time, with framing understood as a multi-dimensional construct involving selection, emphasis, narrative templates, and rhetoric rather than being reducible to topic or sentiment \citep{Otmakhova2024FramingSurveyACL}. Related work quantifies longitudinal divergence such as polarization-driven drift in lexical choice, affect, and semantics \citep{KarjusCuskley2024Polarization}, models deliberation via computational argumentation and interaction dynamics \citep{Irani2025DALiSM}, and uses cross-temporal semantic retrieval over news archives to trace narrative recurrence and transformation \citep{Franklin2024NewsDejaVu}; long-horizon historical corpora further motivate time-aware modeling of cultural change \citep{Hegde2025Chronoberg,Ma2025PeoplesDailyIdeology}.

\subsection{Sparse Autoencoders for Interpretable Representations}
A major interpretability approach decomposes dense neural representations into a small set of latent factors, mitigating multi-feature superposition \citep{Elhage2022Superposition}. Sparse coding and dictionary learning provide classical foundations \citep{OlshausenField1996Sparse,Mairal2010OnlineDictionary}, and sparse autoencoders instantiate these ideas for neural activations\citep{MakhzaniFrey2013kSAE}. In mechanistic interpretability, SAE-style dictionary learning has been used to extract more nearly monosemantic features from transformer activations and enable feature-level analyses at scale \citep{Cunningham2023SAE,Bricken2023Monosemantic,Gao2024ScalingSAE}, often paired with automated labeling and validation pipelines \citep{Bills2023NeuronExplainer,ONeill2024DisentanglingDenseEmbeddings}. By contrast, direct comparisons of word vectors can be affected by anisotropy, rogue dimensions, robustness issues, and social bias \citep{Ethayarajh2019Contextual,TimkeyVanSchijndel2021RogueDimensions,GuoCaliskan2021Bias}, motivating sparse-feature-based analyses for more robust characterization of linguistic and semantic phenomena \citep{Matthews2025KSparseLSC,LinguaLens2025}.

\section{Methodology}

\begin{figure*}[tp]
    \centering
    \includegraphics[width=0.97\textwidth]{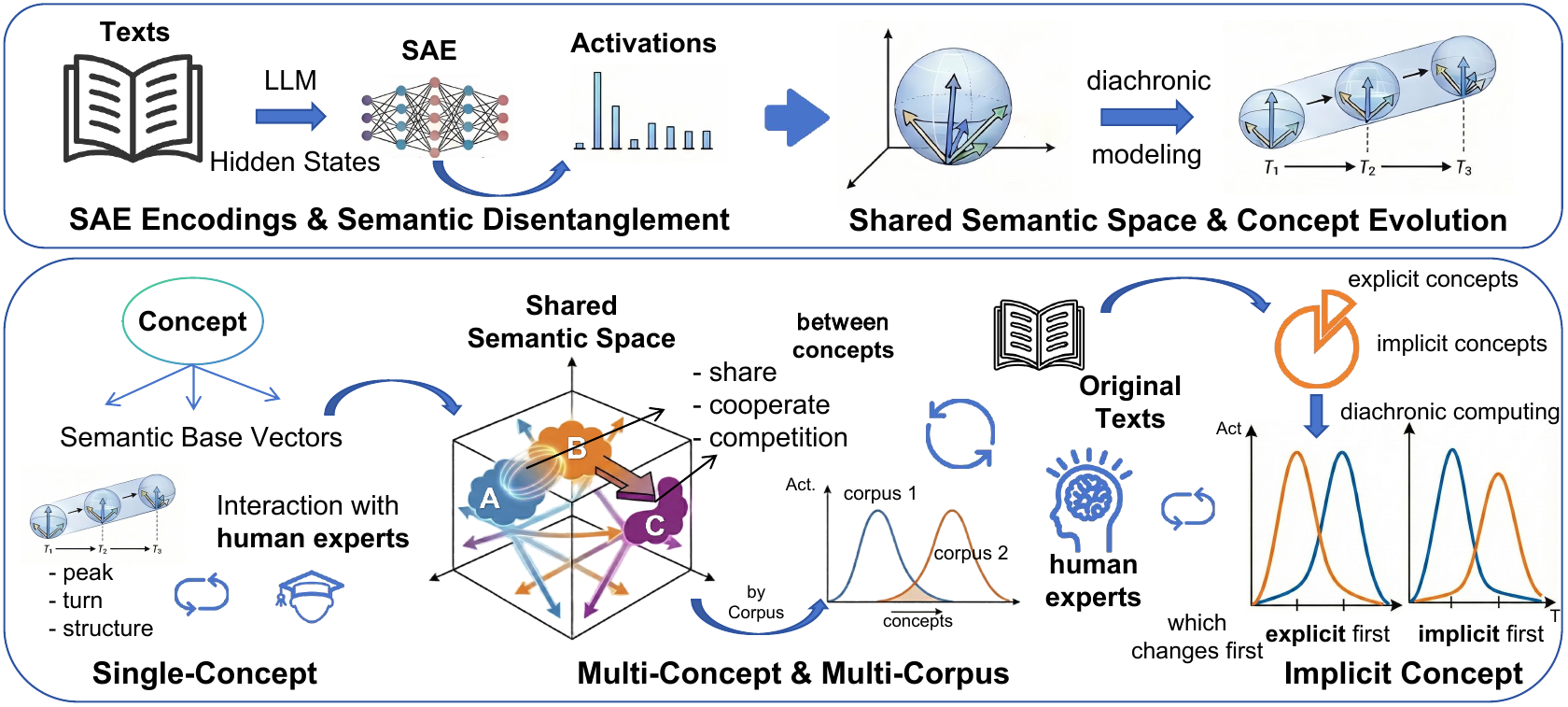}
    \vspace{-0.01in}
    \caption{\textbf{Framework overview}. We encode diachronic texts with an LLM and project hidden states into sparse SAE activations to obtain interpretable semantic base vectors, forming a shared semantic space. We then perform diachronic modeling in this space to quantify concept dynamics (e.g., peak windows, turning points, and structural shifts) for single-concept and multi-concept/multi-corpus settings, and further distinguish implicit concept signals from explicit lexical mentions, offering new insights for the study of intellectual and conceptual evolution.}
    \label{fig:method}
\end{figure*}

\subsection{Sparse Auto-Encoder Representations}
\label{sec:sae}
We use pretrained Sparse Auto-Encoders (SAEs) as a fixed system for diachronic analysis. Given timestamped textual units $\{(x_i,t_i)\}$, where each $x_i$ is a sentence, we encode each $x_i$ with a frozen pretrained LLM to obtain token-level hidden states $\mathbf{h}_{i,j}\in\mathbb{R}^d$. For the main pipeline, $\mathbf{h}_{i,j}$ denotes the residual-stream state at Layer~29 of Llama-3.1-8B-Instruct, and we use the pretrained Layer-29 OpenSAE as a fixed mapping throughout unless otherwise noted; additional technical background is given in Appendix~\ref{app:opensae-background}. A pretrained SAE then maps each hidden state into a sparse feature space,
\begin{equation}
\mathbf{z}_{i,j}=f_{\mathrm{SAE}}(\mathbf{h}_{i,j})\in\mathbb{R}^K,
\end{equation}
and we aggregate token activations into a text-level sparse representation $\mathbf{z}_i=\mathrm{Agg}_j(\mathbf{z}_{i,j})$ using max pooling over tokens. Throughout, we \emph{do not} update parameters of either the LLM or the SAE; we treat them as fixed nonlinear mappings and analyze only the resulting activations.

\paragraph{Selecting drifting base vectors and interpretation.}
\label{sec:method}
For each SAE base vector (dimension) $k$, we summarize its activation by time slice $s$ as $\mu_{k,s}=\mathbb{E}[\mathbf{z}_{i,k}\mid t_i\in s]$, and define its cumulative drift over a given period as
\begin{equation}
D_k=\sum_{s=2}^{S}\left|\mu_{k,s}-\mu_{k,s-1}\right|.
\end{equation}
We then select base vectors with the largest $D_k$ (i.e., maximal total drift). To interpret each selected vector, we retrieve its highest-activating texts and submit these contexts to human experts, who assign semantic descriptions grounded in the evidence provided by the activating passages.

\subsection{Diachronic Analysis}
\label{sec:diachronic}

\method treats conceptual history as the reconfiguration of \emph{interpretable semantic components} across time and discourse fields, operationalized as SAE \emph{base vectors} in a shared sparse activation space (Section~\ref{sec:sae}). Each concept is represented as a structured combination of base-vector activations, enabling scalable distant reading with evidential accountability: quantitative signals locate salient periods and components, and claims are traceable to high-activating textual evidence.

\subsubsection{Constructing the Concept--Corpus Atlas}
We slice each periodical into yearly bins and, for each (concept, corpus) pair, compute a small set of navigational statistics to enable macro characterization and cross-material comparison (formal definitions in Appendix~\ref{app:diachronic-metrics}). The \emph{peak year} is the year where the concept magnitude (slice-aggregated activation over $\mathcal{S}_c$) is maximal. The \emph{turning point} is the year with the strongest adjacent-slice change; its signed intensity $I$ equals the signed magnitude of that change (positive for increase, negative for decrease). The \emph{diversity} $H$ quantifies how dispersed the concept composition is across base vectors, defined as the normalized entropy of base-vector contribution shares (higher $H$ indicates a more diffuse mixture).

\subsubsection{Single-Concept Diachronic Decomposition}
\label{sec:single-concept}
For a single concept $c$, we decompose its diachronic variation in the shared SAE space into a small set of informative base-vector drivers. For each time slice $s$, we compute the mean activation of base vector $k$ as $\mu_{k,s}=\mathbb{E}[z_{i,k}\mid t_i\in s]$, and select a concise set of base vectors with high activation in texts salient for concept $c$ and salient temporal variation. We then track $\mu_{k,s}$ trajectories over time and use adjacent-slice differences (or relative change rates) to locate localized surges and reversals, yielding fine-grained quantitative signals that guide subsequent analysis.

\subsubsection{Multi-Concept Comparability and Interaction}
\label{sec:multi-concept}

Because all concepts are embedded in the same base-vector coordinate system, comparisons across concepts do not require retraining concept-specific spaces. We therefore analyze multi-concept dynamics at the level of (i) orientation trajectories and (ii) shared base-vector signals, which preserves comparability while keeping interpretation tethered to evidence. Methodologically, this supports humanities questions about conceptual co-evolution, alignment, and tension as relations among interpretable components, rather than as opaque movements in embedding space. The corresponding formal quantities used for multi-concept comparison are specified in Appendix~\ref{app:diachronic-metrics}.

\subsubsection{Cross-Corpus Comparability}
\label{sec:cross-corpus}

Let $\mathcal{R}$ denote a set of corpora (e.g., periodicals or genres). Cross-corpus analysis conditions the same conceptual quantities on $r_i\in\mathcal{R}$ and computes corpus-aware summaries within each time slice. To keep heterogeneous corpora comparable, we focus on \emph{salient realizations} of each concept within each corpus: for each $(c,r)$ we select a high-quantile subset $\mathcal{I}_{c,r}$ of text units by concept magnitude and summarize diachronic coverage and composition only within this set (formal definitions in Appendix~\ref{app:diachronic-metrics}). Since all corpora share the same SAE space, we further align \emph{semantic components} across corpora by comparing the overlap of Top-$30$ drifting base vectors, separating shared components from corpus-specific reweighting.

\subsubsection{Implicit Concept Computation and Interpretation}
\label{sec:implicit}

Concepts are often expressed without their canonical names, via stable discursive patterns. We therefore split the salient set $\mathcal{I}_{c,r}$ into \emph{lexically anchored} vs.\ \emph{implicit} contexts by whether canonical lexemes of concept $c$ are present (see Appendix~\ref{app:implicit} for details), and let $\mathcal{I}^{\mathrm{Imp}}_{c,r}$ be the implicit subset. We quantify implicit realization by the ratio
\begin{equation}
\bar r_{c,r}
=
\frac{\sum_{i\in\mathcal{I}^{\mathrm{Imp}}_{c,r}} m_i^{(c)}}
{\sum_{i\in\mathcal{I}_{c,r}} m_i^{(c)}}.
\end{equation}

\paragraph{Humanistic interpretation.}
We treat computational signals as methodological scaffolding and follow a humanistic interpretation protocol based on expert reading of highest-activating contexts (Appendix~\ref{app:interp}).

\section{Experiments}

\subsection{Experiment Setup}

We conduct our experiments with the pretrained OpenSAE family \citep{opensae} attached to Llama-3.1-8B-Instruct \citep{grattafiori2024llama3}. Unless otherwise stated, all main reported results use the Layer-29 SAE over the residual stream; only the dedicated cross-layer robustness analysis reruns the same pipeline with Layers~06/14/22/29. We compile and curate complete runs of several representative periodicals from modern Chinese history, including \emph{New Youth (Xinqingnian)} and \emph{The Guide (Xiangdao)}, digitized from print editions via OCR, totaling \textit{3,277} issues and \textit{8,030,009} characters. We further select four foundational concepts as case studies: \emph{individual}, \emph{society}, \emph{nation}, \emph{world}.

\subsection{Main Results}

\subsubsection{Concept--Corpus Atlas}
\label{sec:atlas-main}
\begin{table}[t]
    \centering
    \setlength{\tabcolsep}{5pt}
    \resizebox{1.0\linewidth}{!}{
    \begin{tabular}{l c c c c}
        \toprule
        \textbf{Concept} & $\boldsymbol{\bar r}$ & $\boldsymbol{H}$ & \textbf{Peak} & \textbf{Turn (year, $I$)} \\
        \midrule
        \multicolumn{5}{l}{\textcolor{gray}{\textbf{\textit{New Youth}}}} \\
        \; \emph{individual} & 0.920 & 0.741 & 1920 & (1918, $+0.226$) \\
        \; \emph{nation}     & 0.921 & 0.743 & 1924 & (1918, $+0.116$) \\
        \; \emph{society}    & 0.595 & 0.368 & 1922 & (1918, $-0.213$) \\
        \; \emph{world}      & 0.900 & 0.683 & 1926 & (1918, $+0.230$) \\
        \midrule
        \multicolumn{5}{l}{\textcolor{gray}{\textbf{\textit{The Guide}}}} \\
        \; \emph{individual} & 0.963 & 0.763 & 1923 & (1923, $+0.0665$) \\
        \; \emph{nation}     & 0.860 & 0.568 & 1926 & (1923, $-0.0810$) \\
        \; \emph{society}    & 0.759 & 0.557 & 1924 & (1924, $+0.152$) \\
        \; \emph{world}      & 0.778 & 0.786 & 1925 & (1926, $-0.0852$) \\
        \bottomrule
    \end{tabular}
    }
    \caption{A slice of the Concept--Corpus Atlas. We report ratio-based and structural signals only: de-lexicalization strength $\bar r$, diversity $H$ (normalized entropy over orientations), and diachronic anchors (Peak year and the strongest turning point). Turn intensity $I$ is signed: $I>0$ indicates an upward shift and $I<0$ a downward shift at the turning year.}
    \label{tab:atlas-slice}
\end{table}

We slice each corpus into time-indexed segments and pass them through the SAE to obtain sparse-feature activations. Following Section~\ref{sec:method}, we compute a compact set of reproducible statistics for each (concept, corpus) pair: de-lexicalization strength $\bar r$, diversity $H$, the peak year, and the strongest turning point (Table~\ref{tab:atlas-slice}).

Overall, de-lexicalized concept practice is substantial, with $\bar r$ ranging from 0.595 to 0.963. In \emph{New Youth}, \emph{individual}, \emph{nation}, and \emph{world} are strongly de-lexicalized ($\bar r \approx 0.900$--$0.921$), whereas \emph{society} is less so ($\bar r=0.595$), indicating more lexically anchored usage. Structural dispersion also varies: \emph{society} in \emph{New Youth} is markedly concentrated ($H=0.368$), while \emph{world} in \emph{The Guide} is more diffuse ($H=0.786$). Diachronic anchors sharply separate the corpora: all four concepts in \emph{New Youth} share their strongest turn at 1918 with larger magnitudes ($|I|=0.116$--$0.230$), whereas the strongest turns in \emph{The Guide} shift later (1923--1926) and are weaker overall (largest $|I|=0.152$). These compact signals provide reproducible anchors for selecting downstream case studies in the subsequent analyses.

\subsubsection{Single-Concept Analysis}

\begin{figure}[tp]
    \centering
    \includegraphics[width=0.98\linewidth]{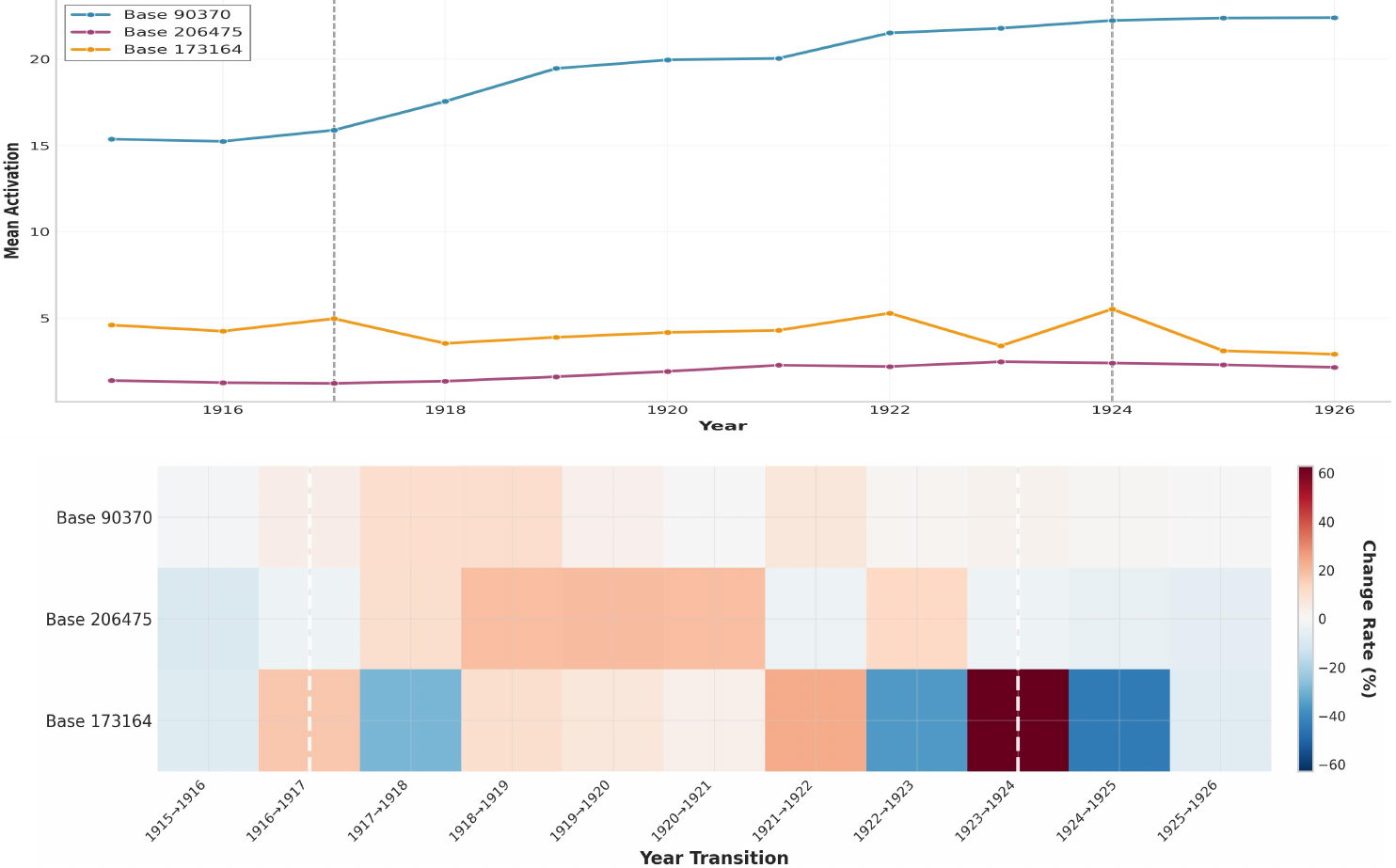}
    \caption{Single-concept decomposition for \emph{individual} in \emph{New Youth}. Top: mean activation trajectories of selected SAE base vectors across years, with dashed markers indicating salient transition windows. Bottom: year-to-year relative change rates for the same base vectors, highlighting localized surges and reversals that serve as navigational anchors for close reading.}
    \label{fig:singlecon-individual}
\end{figure}

Guided by the \emph{Concept--Corpus Atlas}, we zoom in on a single concept, \emph{individual}, in \emph{New Youth} to illustrate the value of SAE-based decomposition: an apparently holistic diachronic trajectory can be factorized into a small number of interpretable semantic drivers. We select several highly activated base vectors with strong explanatory power and track their mean activations across time slices. To reduce cognitive load, we refer to these components below by short semantic labels rather than by raw base indices; Table~\ref{tab:label-index} maps each label to its constituent SAE base(s) and representative bilingual evidence.

As shown in Fig.~\ref{fig:singlecon-individual}, the evolution of \emph{individual} decomposes into three primary semantic strands: \textit{Actorhood} rises steadily over the long run and remains high in later years; \textit{Individualism as Discourse} exhibits a sharp surge in a salient turning window followed by a rapid retreat; and \textit{Property and Economic Individuality} stays comparatively low but increases gradually, contributing as a persistent background dimension. Overall, the concept’s change is driven by differentiated dynamics at the semantic-component level rather than incidental fluctuations in surface frequency or collocational patterns.

This decomposition further indicates that, in this diachronic corpus, \emph{individual} is not a semantically homogeneous object that can be adequately captured by a single scalar indicator; instead, it forms a ``semantic assemblage'' composed of multiple strands that can vary relatively independently (e.g., actorhood, discursive individualism, and property/economic individuality). Accordingly, seemingly paradoxical historical observations---for example, periods in which \emph{individualism} as discourse recedes while discussion of \emph{individual} does not diminish---need not be interpreted as conceptual disappearance or analytic misreading. Rather, they reflect shifts in the relative prominence and coordination among internal strands: conceptual continuity derives from the persistence of the assemblage, whereas conceptual turning points emerge from reorganization within the assemblage. The SAE-based framework thus provides a principled lens for characterizing the internal heterogeneity of concepts and tracing its diachronic evolution.

\subsubsection{Multi-concept analysis}

\begin{figure*}[tp]
    \centering
    \includegraphics[width=0.97\textwidth]{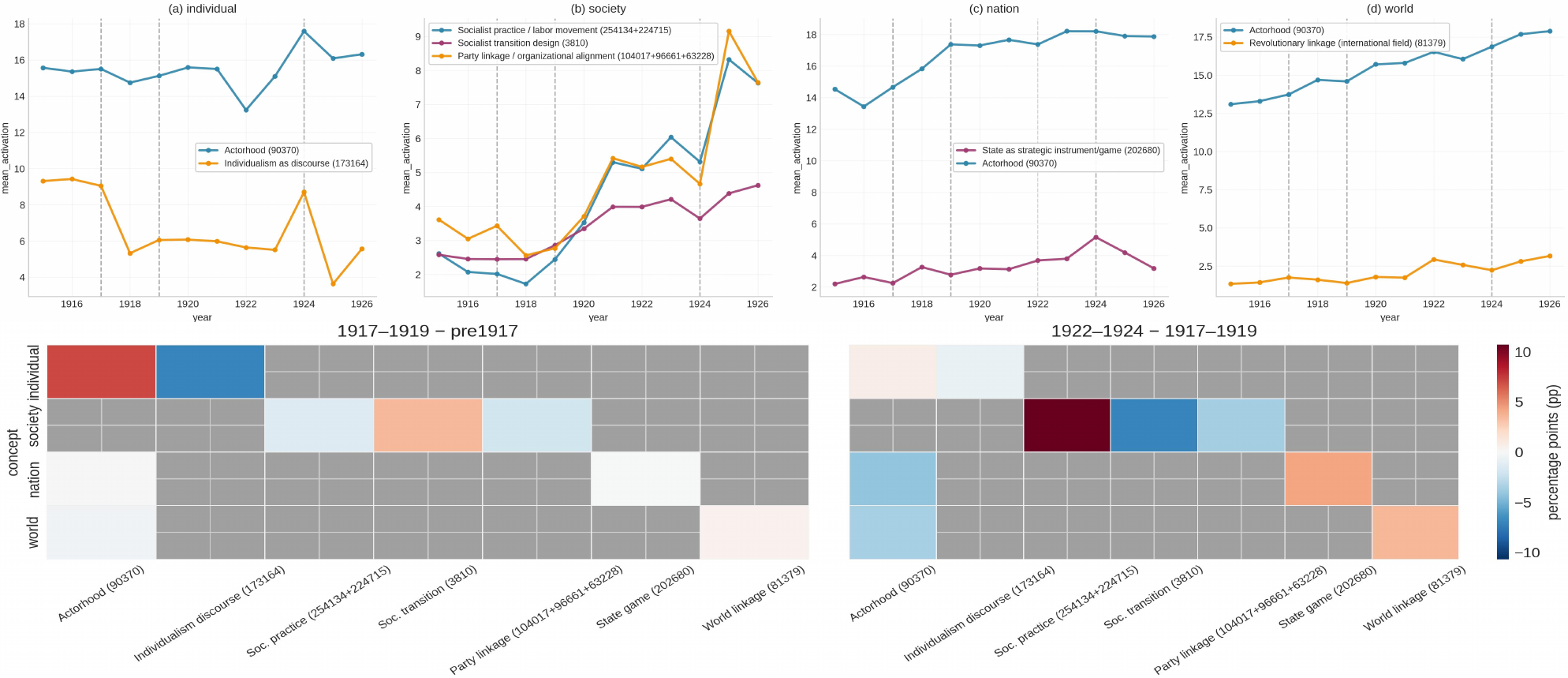}
    \vspace{-0.01in}
    \caption{Multi-concept dynamics in a shared SAE space. Top: for \emph{individual}, \emph{society}, \emph{nation}, and \emph{world}, we track a small set of salient sparse features and plot their mean activations by year, enabling direct comparison in the same semantic coordinate system. Bottom: window-level composition shifts are summarized as difference heatmaps of feature-share changes across two contrasts, $1917$--$1919$ relative to \textit{pre}1917 and $1922$--$1924$ relative to $1917$--$1919$, revealing coordinated within-concept reweighting around historical junctures.}
    \label{fig:multiconcept}
\end{figure*}

We conduct an aligned comparison of four concepts in \emph{New Youth}---\emph{individual}, \emph{society}, \emph{nation}, and \emph{world}---within a shared SAE semantic coordinate system. Concretely, we select a small set of representative sparse features for each concept, compute their mean activations on yearly slices, and aggregate feature shares at the window level. We then use difference heatmaps to contrast within-concept reweighting in $1917$--$1919$ relative to \textit{pre}1917 and in $1922$--$1924$ relative to $1917$--$1919$, thereby characterizing multi-concept dynamics on the same timeline and in the same semantic units. Throughout this discussion, we refer to the selected features by stable semantic labels; Table~\ref{tab:label-index} provides the label-to-base mapping.

As visualized in Fig.~\ref{fig:multiconcept}, the difference patterns indicate structured semantic reconfiguration around major historical junctures. In $1917$--$1919$ relative to \textit{pre}1917, \emph{individual} increases \textit{Actorhood} while decreasing \textit{Individualism as Discourse}; in parallel, \emph{society} exhibits a relative rise in \textit{Societal Transition and Institutional Design}, accompanied by relative declines in \textit{Organized Praxis and Labor-Movement Alignment} and \textit{Party Linkage and Organizational Alignment}. Moving to $1922$--$1924$, reweighting further intensifies: \emph{society} shifts from \textit{Societal Transition and Institutional Design} toward \textit{Organized Praxis and Labor-Movement Alignment}; \emph{world} strengthens \textit{Revolutionary International Field} while \textit{Actorhood} relatively decreases; and \emph{nation} increases \textit{Nation-State as Strategic Instrument}, reflecting a temporally localized restructuring of the national frame.

These shifts suggest that the central object of concept computation is not the mere appearance or disappearance of a concept, but the reweighting of its internal semantic components under changing historical pressures, which in turn reshapes its discursive function and historical role. \emph{Individual} becomes more strongly organized around narratives of agentive position and responsibility; \emph{society} reallocates emphasis between institutional blueprints and organized praxis; \emph{world} is more tightly embedded in transnational revolutionary linkages; and \emph{nation} is foregrounded as a strategic instrument and a structure of contestation. Building on this view, we operationalize the otherwise hard-to-systematize notion of ``shifts in semantic emphasis'' as fine-grained reweighting and cross-concept alignment in a shared SAE space, opening up additional quantitative entry points while preserving the nuance of humanities interpretation.

\subsubsection{Cross-corpus Analysis}

\begin{figure}[t]
    \centering
    \includegraphics[width=0.97\linewidth]{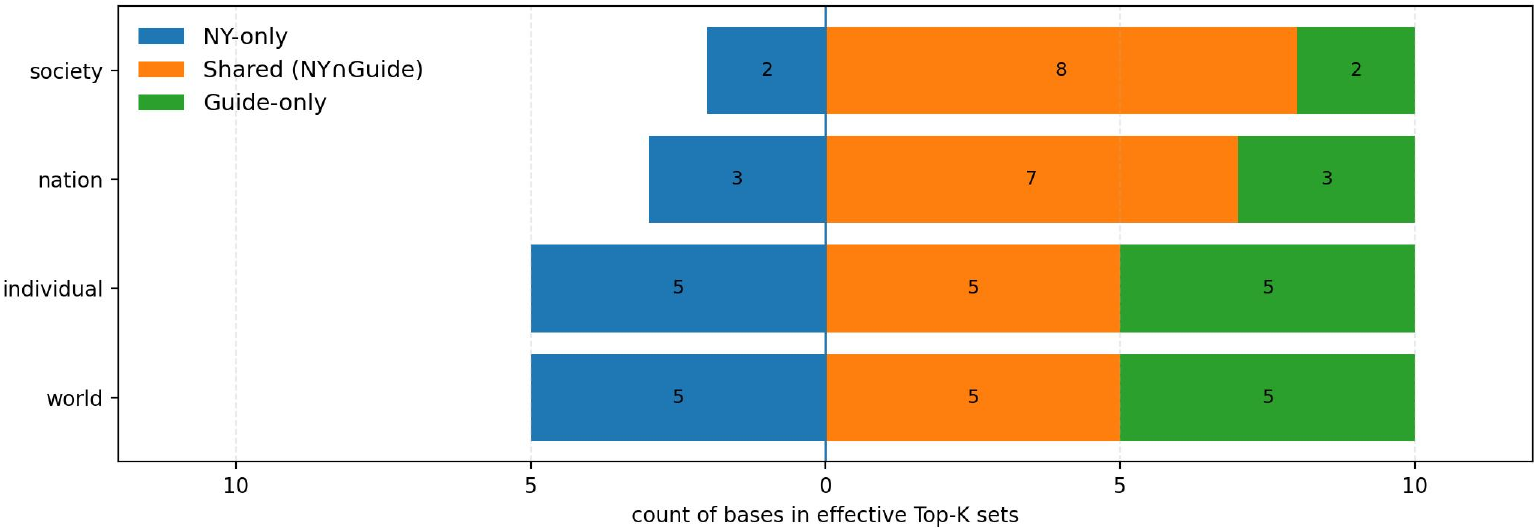}
    \vspace{-0.08in}
    \caption{Cross-corpus decomposition of effective Top-$30$ drifting bases in a shared SAE space. For each concept, we partition the drifting bases into \emph{New Youth}-only, shared (\emph{New Youth} $\cap$ \emph{The Guide}), and \emph{The Guide}-only components, visualizing how cross-corpus overlap versus corpus-specific components vary across concepts.}
    \label{fig:multicorpus-overlap}
    \vspace{-0.08in}
\end{figure}

We further conduct a cross-corpus comparison between \emph{New Youth} and \emph{The Guide}. Within the same SAE semantic coordinate system, we first quantify the overlap of the Top-$30$ drifting bases to distinguish a cross-corpus \emph{stable semantic skeleton} from \emph{corpus-specific semantic reweighting}. We then use, for each base, its full-range top-$30$ highest-activating sentences as the evidence pool and display $8$ representative sentences for close reading, attributing divergences to concrete discourse mechanisms (e.g., genre-specific communicative tasks, polemical targets, and mobilization-oriented rhetoric). This workflow makes cross-material comparison less dependent on ad hoc sentence selection: instead, comparable semantic components provide a reproducible reading guide, enabling more transparent and falsifiable historical interpretation.

We illustrate the value of this multi-source perspective with the concept \emph{world}. Its relatively low overlap in Top-$30$ drifting bases suggests that \emph{world} is not organized by an identical set of semantic handles across the two corpora; nevertheless, the shared drifting bases still provide a common baseline, consistently anchoring \emph{world} to a macro-structure of revolution, class struggle, and imperialism. The humanities-oriented gain lies in the \emph{localizable} nature of the differences: in \emph{New Youth}, \emph{world} more readily intertwines with intellectual debate, epistemic framing, and cultural critique, forming a world-view oriented toward conceptual renovation; in \emph{The Guide}, \emph{world} is more often compressed into a field of camp alignment, organizational mobilization, and institutional claims, yielding a sharper action orientation and political demarcation. In this sense, cross-corpus comparison turns the heterogeneity of \emph{world} into a structural finding: the same-named concept is reweighted by different discursive tasks on top of a shared backbone, resulting in distinct rhetorical foci and historical functions. Our innovation is to operationalize cross-corpus comparability as a decomposition into shared versus corpus-specific semantic components with aligned evidence, thereby combining the nuance of close reading with a scalable and reproducible computational procedure.

\subsubsection{Implicit Concept Computation}

\begin{figure}[tp]
    \centering
    \includegraphics[width=0.98\linewidth]{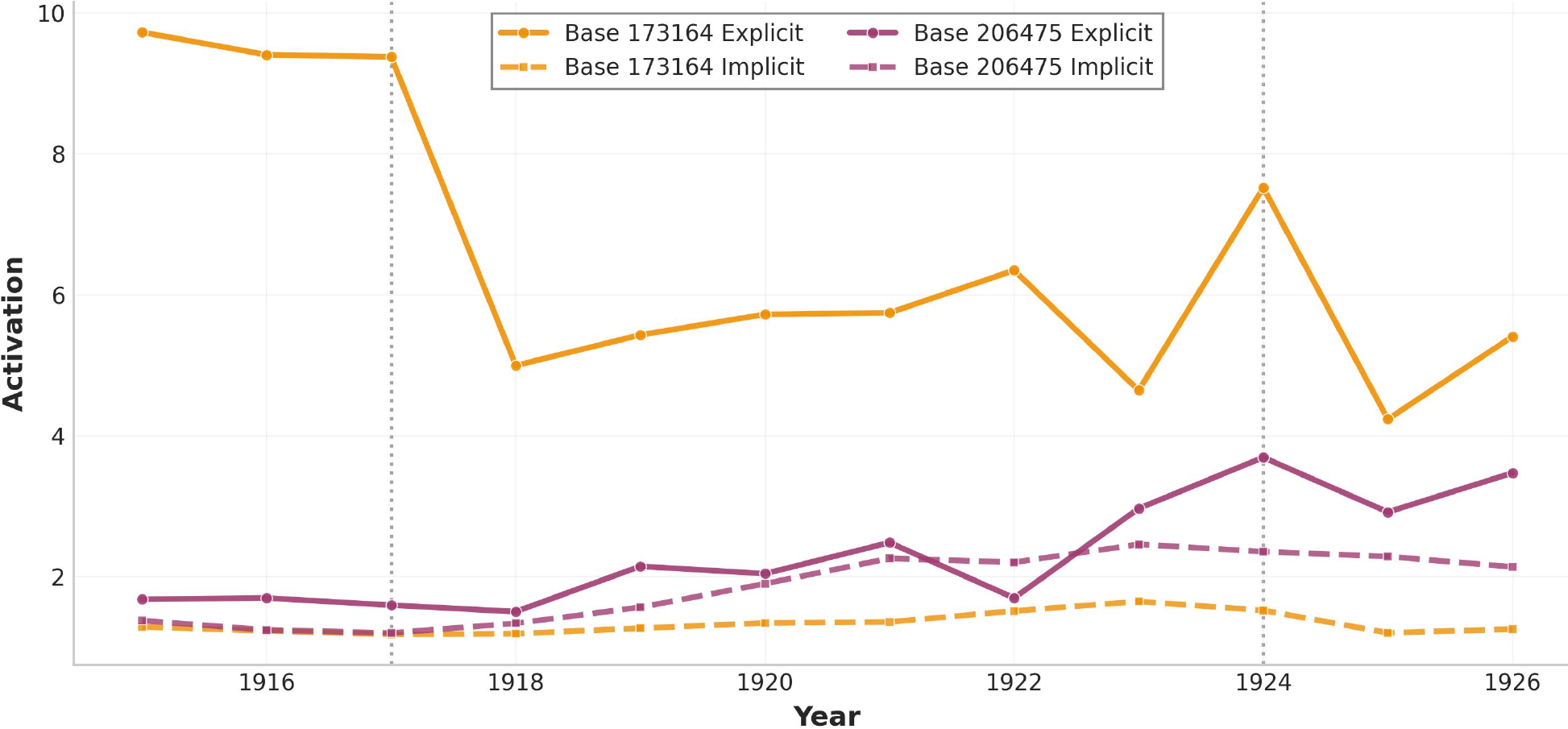}
    \caption{Explicit vs.\ implicit concept computation for \emph{individual} in \emph{New Youth}. Yearly mean activations are shown for \textit{Individualism as Discourse} and \textit{Property and Economic Individuality} in anchor-present (Explicit) and anchor-absent (Implicit) contexts. Vertical dashed lines indicate salient transition windows.}
    \label{fig:implicit-individual}
\end{figure}
We further contrast \emph{Explicit} (anchor-present) and \emph{Implicit} (anchor-absent) evidence for \emph{individual} in \emph{New Youth} by tracking yearly mean activations of two representative semantic components, \textit{Individualism as Discourse} and \textit{Property and Economic Individuality} (Fig.~\ref{fig:implicit-individual}). The results show that concept-related semantics remains substantial even when lexical anchors are absent, and that implicit dynamics are not a trivial mirror of explicit trajectories, providing additional and reproducible navigational signals for close reading.
A separate historian validation on $20$ sampled anchor-absent high-activation cases per concept yields semantic-consistency rates of 90\% for \emph{society}, 100\% for \emph{nation}, 100\% for \emph{world}, and 75\% for \emph{individual}; representative examples are reported in Appendix~\ref{app:implicit-validation}.

For \textit{Individualism as Discourse}, the Explicit trajectory is high in 1915--1917, drops sharply in 1918, and rebounds during 1922--1924 with a local peak around 1924. In contrast, the Implicit trajectory stays lower and smoother, reaching a relative high around 1923 before retreating. This divergence suggests that explicit articulation is more sensitive to episodic contestation and rhetorical intensity, whereas anchor-absent semantic practice is more stable, indicating persistence of the concept beyond overt lexical naming.

For \textit{Property and Economic Individuality}, Implicit evidence rises steadily from 1919 and forms a plateau around 1922--1923, while the Explicit curve exhibits a pronounced jump only in 1923--1924 and peaks around 1924, displaying an implicit-leads-explicit pattern. This implies that institutional and economic semantic components may diffuse via stable semantic carriers across heterogeneous topics before becoming explicitly foregrounded and lexically consolidated in more concentrated political and institutional debates.

Overall, Explicit signals are closer to public naming and the intensity of contestation, whereas Implicit concept computation better captures semantic permeation and cross-topic diffusion. Their divergence and lead--lag relations serve as key criteria for locating turning windows for downstream close reading. This explicit--implicit contrastive quantification framework renders the \emph{presence and change of a concept even when its keyword is not explicitly attested in the text} into computable and interpretable outcomes, advancing the traditionally intuition-driven identification of unspoken ideas into a systematic line of inquiry.

\subsubsection{Cross-layer Analysis}
\begin{table}[t]
\centering
\small
\setlength{\tabcolsep}{6pt}
\begin{tabular}{llccc}
\toprule
Concept & Layer & Peak & Turn & Avg. Jaccard \\
\midrule
\multirow{4}{*}{\emph{individual}} 
 & 06 & 1923 & 1918 & 0.50 \\
 & 14 & 1923 & 1918 & 0.42 \\
 & 22 & 1920 & 1918 & 0.48 \\
 & 29 & 1920 & 1918 & 0.33 \\
\midrule
\multirow{4}{*}{\emph{nation}} 
 & 06 & 1924 & 1918 & 0.57 \\
 & 14 & 1924 & 1918 & 0.53 \\
 & 22 & 1924 & 1918 & 0.60 \\
 & 29 & 1924 & 1918 & 0.57 \\
\bottomrule
\end{tabular}
\caption{Cross-layer robustness results on \emph{New Youth}. ``Avg.\ 2-gram Jaccard'' is the mean 2-gram Jaccard similarity between the evidence-context fingerprint of a given layer and those of the other three layers.}
\label{tab:layer-robust}
\end{table}

To test whether our diachronic signals depend on the SAE layer, we rerun the same pipeline on the same corpus while only switching SAE layers (06/14/22/29), and use \emph{individual} and \emph{nation} in \emph{New Youth} as representative concepts. For each layer, we report (i) the \emph{peak year}, defined as the year in which the concept magnitude attains its maximum; (ii) the \emph{turning year}, defined as the year corresponding to the strongest adjacent-slice change; and (iii) \emph{Avg.\ Jaccard}, the mean 2-gram Jaccard similarity between the evidence-context fingerprint of a given layer and those of the other three layers (Table~\ref{tab:layer-robust}). As shown in Table~\ref{tab:layer-robust}, the turning year is identical across all four layers for both concepts (1918 throughout), indicating that the turning-point signal used to locate major semantic reconfiguration is insensitive to layer choice and can serve as a reproducible temporal anchor.

Meanwhile, peak-year localization exhibits concept-dependent layer sensitivity. For \emph{nation}, the peak year is fully stable (1924 across all layers), suggesting that the intensity-based localization of \emph{nation} is robust to representational depth in our setup. In contrast, \emph{individual} shows a consistent stratification: lower layers (06/14) peak in 1923, whereas higher layers (22/29) peak earlier in 1920. This pattern suggests that, while the major turning structure is preserved across layers, different representational depths may emphasize different facets of the concept and thus shift where the maximal concentration of evidence is localized.

This layer effect is also reflected in contextual consistency. Using evidence-context 2-gram fingerprints, \emph{nation} exhibits higher and relatively stable cross-layer agreement (Avg.\ Jaccard $\approx 0.53$--$0.60$), whereas \emph{individual} is less consistent overall and becomes notably more divergent at the highest layer (Layer~29: Avg.\ Jaccard $=0.33$). Overall, the cross-layer experiment supports a robust diachronic backbone (the 1918 turning point) while indicating that peak localization and the concrete evidential contexts can vary more for \emph{individual} than for \emph{nation}, providing complementary views on how conceptual signals are distributed across representational depth.

\section{Conclusion}
We propose \method, a unified computational framework for conceptual history: within a shared, alignable SAE semantic space, it jointly characterizes and compares the diachronic dynamics of multiple concepts across heterogeneous sources. We further introduce \emph{implicit concept computation}, which captures stable semantic signals even when a concept is not explicitly stated by its canonical keywords. Overall, \method translates conceptual change in historical texts into an interpretable, comparable, and traceable quantitative structure, offering the humanities and social sciences new tools and perspectives for diachronic text analysis.


\section{Limitations}

Our work has limitations in experimental scale and in the monosemy (semantic univocality) of SAE base vectors.

\paragraph{Experimental scale.}
Because manual curation and interpretation of both corpora and model outputs is labor-intensive, we conduct case studies on only a limited set of periodicals and a limited set of concepts. In principle, the proposed framework has clear potential to scale to much larger collections. Future work may leverage LLM-based agents to assist analysis and synthesis; however, we remain cautious about the reliability of conclusions produced by current LLMs when applied to large-scale humanities and social-science analysis.

\paragraph{Monosemy of base vectors.}
Due to training-cost constraints, the SAE base vectors we obtain are not uniformly monosemous with a clear semantic referent. In our experiments, we observe that some base vectors do not exhibit a stable single meaning, and we exclude them from downstream analysis. Addressing monosemy in practical SAE deployments remains a major open challenge for the field.

\section{Ethical Considerations}
This section discusses the ethical considerations and broader impact of this work:

\paragraph{Potential Risks:}
Our framework provides structured access to latent semantic features and their diachronic dynamics. While the goal is scholarly analysis and evidence-grounded interpretation, similar tooling could be repurposed to engineer persuasive narratives or selectively amplify particular framings in downstream generation settings. To mitigate misuse, we will release code and documentation with clear intended-use statements, emphasize analysis over manipulation, and provide reproducible evaluation scripts so that claims can be independently verified.

\paragraph{Intellectual Property:}
The models and toolkits used in this study, including Llama-3.1-8B(-Instruct) and the OpenSAE framework, are open-source and used for scientific research in accordance with their respective licenses. We will also document provenance for all corpora and derived artifacts, and follow any redistribution constraints associated with the underlying sources.

\paragraph{Data Privacy:}
Our experiments use historical periodicals from the public domain or with established research access. The data does not target individuals and is not intended to contain personal or private information. We additionally apply basic filtering to remove incidental personal identifiers that may appear in historical texts where applicable.

\paragraph{Intended Use:}
\method is intended as a research tool for computational social science and digital humanities: it supports scalable concept analysis, facilitates evidence retrieval for close reading, and enables reproducible measurement of conceptual reconfiguration over time and across sources. It is not intended for decision-making about individuals or for high-stakes profiling.

\paragraph{Documentation of Artifacts:}
We will comprehensively document all released artifacts (corpora metadata, preprocessing, slicing procedures, probe construction, and evaluation metrics), including their domains, time spans, languages, and known limitations, to ensure transparency and reproducibility.

\paragraph{AI Assistants in Research or Writing:}
We employ Cursor for code development assistance and use GPT-5.2 to refine and polish the language of the manuscript.

\section*{Acknowledgments}
This work is supported by NSFC Project 62476009, the Open Project Fund of the State Key Laboratory of Multimedia Information Processing (Project No. SKLMIP-KF-2025-01), and the Tsinghua University Disruptive Innovation Talent Development Program.

\bibliography{1.custom}

\clearpage
\appendix

\section{OpenSAE Technical Background}
\label{app:opensae-background}

\paragraph{Architecture and placement.}
OpenSAE \citep{opensae} is a family of pretrained sparse autoencoders released for LLaMA-3.1-8B. The released SAEs are pretrained on the residual stream, use a context length of $4096$, are trained on $22$B tokens, and expand the LLaMA-3.1-8B hidden state by a factor of $64\times$ into a feature space of size $262{,}144$. In this paper, unless otherwise noted, the main pipeline uses the checkpoint attached to Layer~29. Accordingly, we write $\mathbf{h}^{(29)}_{i,j}\in\mathbb{R}^{d}$ for the Layer-29 residual-stream state of token $j$ in text unit $i$.

\paragraph{Encoder, sparse activation, and decoder.}
OpenSAE maps each layer-specific hidden state to a high-dimensional sparse feature space. Let $\tilde{\mathbf{h}}^{(29)}_{i,j}$ denote the normalized input passed to the encoder. The dense pre-activation vector is
\begin{equation}
\mathbf{a}_{i,j}
=
W_{\mathrm{enc}}\tilde{\mathbf{h}}^{(29)}_{i,j}
+ \mathbf{b}_{\mathrm{enc}}
\in\mathbb{R}^{K}.
\label{eq:opensae-enc}
\end{equation}
Here $K=262{,}144$ is the total feature dimension of the released OpenSAE checkpoint family used in our study.
OpenSAE then applies a TopK sparsification operator that retains only a small set of activated coordinates:
\begin{equation}
\left[\mathbf{z}_{i,j}\right]_m
=
\begin{cases}
\left[\mathbf{a}_{i,j}\right]_m,
& m\in \mathrm{TopK}_{\kappa}(\mathbf{a}_{i,j}),\\
0,
& \text{otherwise},
\end{cases}
\label{eq:opensae-topk}
\end{equation}
where $\mathrm{TopK}_{\kappa}(\mathbf{a}_{i,j})$ denotes the index set of the retained features and $\kappa$ is the number of active features kept for each token. The decoder reconstructs the residual-stream state as
\begin{equation}
\hat{\mathbf{h}}^{(29)}_{i,j}
=
W_{\mathrm{dec}}\mathbf{z}_{i,j}
+ \mathbf{b}_{\mathrm{dec}}.
\label{eq:opensae-dec}
\end{equation}
The sparse activation vector $\mathbf{z}_{i,j}$ is the object used throughout our diachronic pipeline.

\paragraph{How the SAE outputs are interpreted.}
The primary output used in our analysis is the sparse activation vector $\mathbf{z}_{i,j}$. An active coordinate $[\mathbf{z}_{i,j}]_m>0$ indicates that the token-level residual state $\mathbf{h}^{(29)}_{i,j}$ aligns with feature $m$ strongly enough to survive the TopK selection step, while its magnitude indicates the relative strength of that alignment for the given token. We therefore interpret the nonzero support of $\mathbf{z}_{i,j}$ as a sparse inventory of semantic components expressed by the token in context. The associated decoder direction, given by the $m$-th column of $W_{\mathrm{dec}}$ up to implementation convention, specifies how that feature contributes back to the reconstructed hidden state and thus provides the geometric direction that the feature adds to the residual stream.

This interpretation is evidential rather than purely label-based. A feature is not treated as meaningful merely because it has a nonzero index; instead, its interpretation is established by inspecting its highest-activating contexts, verifying that those contexts exhibit a stable semantic regularity, and then assigning a description grounded in that regularity. At the text level, aggregated activations $\mathbf{z}_i$ indicate the relative prevalence of interpretable semantic components within a sentence, rather than the presence of a single categorical meaning. In the main pipeline, $\mathbf{z}_i$ is obtained by max pooling token-level SAE activations within the sentence. By contrast, the reconstruction $\hat{\mathbf{h}}^{(29)}_{i,j}$ is not itself read as a human-interpretable semantic object; it is used as evidence that the retained sparse features preserve substantial information about the original residual-stream state.

\paragraph{Training objective.}
At the level documented by the public OpenSAE interface, the training loss can be written as
\begin{align}
\mathcal{L}_{\mathrm{SAE}}
&=
\lambda_{\mathrm{rec}}\mathcal{L}_{\mathrm{rec}}
+ \lambda_{\mathrm{AuxK}}\mathcal{L}_{\mathrm{AuxK}}
\nonumber\\
&\quad
+ \lambda_{\mathrm{MTK}}\mathcal{L}_{\mathrm{MultiTopK}}
+ \lambda_{1}\mathcal{L}_{1},
\label{eq:opensae-loss}
\end{align}
where the reconstruction term is the squared $L_2$ loss between the input hidden state and its reconstruction,
\begin{equation}
\mathcal{L}_{\mathrm{rec}}
=
\mathbb{E}_{i,j}
\left[
\left\|
\hat{\mathbf{h}}^{(29)}_{i,j}
- \mathbf{h}^{(29)}_{i,j}
\right\|_2^2
\right].
\label{eq:opensae-rec}
\end{equation}
The remaining terms correspond to the AuxK loss, the Multi-TopK loss, and the $L_1$ regularization term exposed by OpenSAE \citep{opensae}. Since our study uses released pretrained checkpoints rather than retraining the SAE, these auxiliary terms matter only as part of the pretrained representation family from which our sparse activations are drawn.

\paragraph{Interpretive role in this paper.}
In our use case, both the LLM and the SAE remain frozen. We do not optimize the SAE further and do not alter its training objective. Instead, we treat the learned sparse features and their associated decoder directions as reusable semantic components, and conduct all downstream analysis on the activations $\mathbf{z}_{i,j}$ and their text-level aggregations $\mathbf{z}_i$. This fixed-feature regime is what enables the subsequent analyses of drift, cross-corpus overlap, implicit realization, and evidence-grounded interpretation.

\section{Diachronic Computations and Definitions}
\label{app:diachronic-metrics}

\subsection{Notation and Time Slicing}
\label{app:notation}

\paragraph{Time slicing and conditioning.}
Let $t\in\{1,\dots,T\}$ index ordered time slices. We write $r\in\mathcal{R}$ for corpora/sources and $i$ for sentence-level text units with metadata $(t_i,r_i)$ and SAE activations $\mathbf{z}_i$.

\paragraph{Concept-level quantities and layer convention.}
Unless otherwise noted, all quantities in this appendix are computed from the Layer~29 residual-stream OpenSAE used in the main pipeline, so $\mathbf{z}_i$ denotes the sentence-level max-pooled aggregation of Layer-29 token activations. For each concept $c$, we instantiate a fixed, study-specific set $\mathcal{S}_c$ of expert-validated SAE bases (or base clusters). $\mathcal{S}_c$ is treated as an operational concept definition for downstream diachronic computation rather than as the output of a separate automatic learning algorithm. For a text unit $i$ and component $s\in\mathcal{S}_c$, let $Z_i^{(c,s)}\ge 0$ denote the aggregated activation assigned to $s$. We define the concept magnitude of text unit $i$ by
\begin{equation}
m_i^{(c)}
=
\sum_{s\in\mathcal{S}_c} Z_i^{(c,s)}.
\label{eq:concept-magnitude}
\end{equation}

\subsection{Salient Set Construction}
\label{app:salient-set}

\paragraph{Within-corpus salience threshold.}
For each $(c,r)$, define a high-quantile threshold over concept magnitudes within corpus $r$:
\begin{equation}
\tau_{c,r}
=
\mathrm{Quantile}_{q}\!\left(\{\, m_i^{(c)} \mid r_i=r \,\}\right),
\label{eq:tau-cr}
\end{equation}
where $q=0.95$ in all reported experiments.

\paragraph{Salient realizations.}
The salient set used for corpus-aware diachronic summaries is
\begin{equation}
\mathcal{I}_{c,r}
=
\{\, i : r_i=r,\; m_i^{(c)}\ge \tau_{c,r}\,\}.
\label{eq:salient-set}
\end{equation}

\subsection{Aggregation and Composition}
\label{app:aggregation}

\paragraph{Within-corpus, within-time aggregation.}
Define within-corpus, within-time means on the salient set:
\begin{align}
\mu_{c,s,r,t}
&=
\mathbb{E}\!\left[
Z_i^{(c,s)}
\;\middle|\;
i\in\mathcal{I}_{c,r},\; t_i=t
\right],
\label{eq:mu-cssrt}\\
A_{c,r,t}
&=
\mathbb{E}\!\left[
m_i^{(c)}
\;\middle|\;
i\in\mathcal{I}_{c,r},\; t_i=t
\right].
\label{eq:A-crt}
\end{align}

\paragraph{Orientation shares (composition).}
Orientation shares are computed as
\begin{equation}
p_{c,s,r,t}
=
\frac{\mu_{c,s,r,t}}
{\sum_{s'\in\mathcal{S}_c}\mu_{c,s',r,t}+\varepsilon},
\qquad \varepsilon>0,
\label{eq:share-cssrt}
\end{equation}
and we write $\mathbf{p}_{c,r,t}=\{p_{c,s,r,t}\}_{s\in\mathcal{S}_c}$.

\paragraph{Diversity (normalized entropy).}
We quantify the dispersion of semantic composition by normalized entropy:
\begin{equation}
H_{c,r,t}
=
\frac{-\sum_{s\in\mathcal{S}_c} p_{c,s,r,t}\log p_{c,s,r,t}}
{\log |\mathcal{S}_c|}.
\label{eq:diversity-H}
\end{equation}

\subsection{Peak Year, Turning Point, and Reorganization}
\label{app:peaks-turns}

\paragraph{Peak year.}
The peak year is defined as the time slice maximizing concept magnitude:
\begin{equation}
t^{\mathrm{peak}}_{c,r}
=
\arg\max_{t\in\{1,\dots,T\}} A_{c,r,t}.
\label{eq:peak-year}
\end{equation}

\paragraph{Turning point and signed intensity.}
We define the turning point as the time slice with the strongest adjacent-slice change in magnitude:
\begin{equation}
t^{\mathrm{turn}}_{c,r}
=
\arg\max_{t\in\{2,\dots,T\}}
\left|A_{c,r,t}-A_{c,r,t-1}\right|,
\label{eq:turn-year}
\end{equation}
and its signed intensity as
\begin{equation}
I_{c,r}
=
A_{c,r,t^{\mathrm{turn}}_{c,r}}
-
A_{c,r,t^{\mathrm{turn}}_{c,r}-1}.
\label{eq:turn-intensity-signed}
\end{equation}

\paragraph{Structural reorganization of composition.}
We quantify adjacent-slice reorganization of orientation shares by the $L_1$ change:
\begin{align}
\Delta_{c,r,t}
&=
\left\|
\mathbf{p}_{c,r,t}-\mathbf{p}_{c,r,t-1}
\right\|_{1}
\nonumber\\
&=
\sum_{s\in\mathcal{S}_c}
\left| p_{c,s,r,t}-p_{c,s,r,t-1}\right|.
\label{eq:reorg-L1}
\end{align}

\subsection{Implicit Concept Computation}
\label{app:implicit}

\paragraph{Implicit vs.\ anchored subsets.}
Let $\mathcal{I}^{\mathrm{Imp}}_{c,r}$ denote salient implicit contexts and $\mathcal{I}^{\mathrm{Anch}}_{c,r}$ salient lexically anchored contexts, where the partition is determined by whether canonical lexemes of concept $c$ are present.

\paragraph{Implicit-realization ratio.}
The implicit-realization ratio is
\begin{equation}
\bar r_{c,r}
=
\frac{\sum_{i\in\mathcal{I}^{\mathrm{Imp}}_{c,r}} m_i^{(c)}}
{\sum_{i\in\mathcal{I}_{c,r}} m_i^{(c)}}.
\label{eq:implicit-ratio}
\end{equation}

\subsection{Selecting Drifting Base Vectors}
\label{app:drift-bases}

\paragraph{Time-slice mean activation.}
For a base vector $k$ (SAE dimension), define its time-slice mean activation within a specified conditioning set $\mathcal{J}$:
\begin{equation}
\mu_{k,t}=\mathbb{E}[\, z_{i,k}\mid i\in\mathcal{J},\; t_i=t \,].
\label{eq:mu-kt}
\end{equation}

\paragraph{Cumulative drift.}
We define cumulative drift as
\begin{equation}
D_k=\sum_{t=2}^{T}\left|\mu_{k,t}-\mu_{k,t-1}\right|.
\label{eq:drift-Dk}
\end{equation}
We select base vectors with the largest $D_k$ for interpretation. For diachronic evidence, each selected base is localized to the adjacent year pair $[y_1,y_2]$ where $\left|\mu_{k,t}-\mu_{k,t-1}\right|$ is maximal, and we retrieve the top-$5$ highest-activating sentences from $y_1$ and the top-$5$ highest-activating sentences from $y_2$.

\subsection{Cross-Corpus Component Overlap}
\label{app:cross-corpus-formal}

\paragraph{Corpus-conditioned drifting bases.}
For cross-corpus comparison, we use the salient set as the conditioning set, i.e., $\mathcal{J}=\mathcal{I}_{c,r}$, and compute $D_k^{(c,r)}$ by applying Eq.~\eqref{eq:mu-kt}--\eqref{eq:drift-Dk} within each corpus $r$.

\paragraph{Top-$K$ drifting set and Jaccard@K.}
Let $\mathcal{T}^{K}_{c,r}$ denote the set of the $K$ base vectors with the largest $D_k^{(c,r)}$, with $K=30$ in all reported experiments.
For two corpora $r$ and $r'$, we measure overlap as
\begin{equation}
\mathrm{Jaccard@K}(c; r,r')
=
\frac{|\mathcal{T}^{K}_{c,r}\cap \mathcal{T}^{K}_{c,r'}|}
{|\mathcal{T}^{K}_{c,r}\cup \mathcal{T}^{K}_{c,r'}|}.
\label{eq:jaccard-at-k}
\end{equation}
This induces a decomposition into shared components $\mathcal{T}^{K}_{c,r}\cap \mathcal{T}^{K}_{c,r'}$ and corpus-specific components via set differences.

\subsection{Cross-Layer Robustness Metrics}
\label{app:cross-layer-formal}

\paragraph{Layer-conditioned reruns.}
Let $\ell\in\mathcal{L}$ index SAE layers used in robustness tests (e.g., $\mathcal{L}=\{06,14,22,29\}$). For each $\ell$, we rerun the full pipeline with only the SAE layer switched, producing layer-specific trajectories $A^{(\ell)}_{c,r,t}$ and $\mathbf{p}^{(\ell)}_{c,r,t}$.

\paragraph{Peak and turning years across layers.}
Layer-specific peak and turning years are computed by applying Eq.~\eqref{eq:peak-year} and Eq.~\eqref{eq:turn-year} (and intensity by Eq.~\eqref{eq:turn-intensity-signed}) to $A^{(\ell)}_{c,r,t}$.

\paragraph{Evidence-context 2-gram fingerprint and Avg.\ Jaccard.}
Let $\mathcal{E}^{(\ell)}_{c,r}$ be the collection of evidence contexts retrieved for interpretation under layer $\ell$, using for each selected drifting base the peak adjacent year pair and the top-$5$ highest-activating sentences from each of the two years. Define the layer-specific fingerprint as the set of character 2-grams extracted from $\mathcal{E}^{(\ell)}_{c,r}$:
\begin{equation}
\mathcal{F}^{(\ell)}_{c,r}
=
\mathrm{2GramSet}\!\left(\mathcal{E}^{(\ell)}_{c,r}\right).
\label{eq:fingerprint}
\end{equation}
For two layers $\ell$ and $\ell'$, the 2-gram Jaccard similarity is
\begin{equation}
J\!\left(\ell,\ell';c,r\right)
=
\frac{|\mathcal{F}^{(\ell)}_{c,r}\cap \mathcal{F}^{(\ell')}_{c,r}|}
{|\mathcal{F}^{(\ell)}_{c,r}\cup \mathcal{F}^{(\ell')}_{c,r}|}.
\label{eq:jaccard-2gram}
\end{equation}
We report the average cross-layer agreement for layer $\ell$ as
\begin{equation}
\mathrm{AvgJaccard}\!\left(\ell;c,r\right)
=
\frac{1}{|\mathcal{L}|-1}
\sum_{\substack{\ell'\in\mathcal{L}\\ \ell'\ne \ell}}
J\!\left(\ell,\ell';c,r\right).
\label{eq:avg-jaccard}
\end{equation}

\subsection{Humanistic Interpretation Protocol}
\label{app:interp}

\paragraph{Methodological stance.}
We treat computational signals as methodological scaffolding rather than self-sufficient explanations.

\paragraph{Context retrieval.}
For selected orientations and drifting base vectors, we retrieve highest-activating evidence contexts with task-specific rules. For diachronic interpretation, we use each selected base's peak adjacent year pair and retrieve the top-$5$ highest-activating sentences from each year. For cross-corpus interpretation, we use the full-range top-$30$ highest-activating sentences for each base as the evidence pool and display $8$ representative sentences per base. Concretely, we use the retrieved evidence contexts $\mathcal{E}_{c,r}$ as the primary material for interpretation.

\paragraph{Expert annotation and discursive roles.}
Domain experts assign semantic descriptions and articulate discursive roles grounded in the retrieved contexts. To prioritize base vectors most informative for diachronic reading, we select those with maximal cumulative drift (Eq.~\eqref{eq:drift-Dk}) over the period of interest and interpret them through expert analysis of their activating texts.

\section{Historian Validation of Implicit Concept Computation}
\label{app:implicit-validation}

\paragraph{Validation setup.}
To test whether anchor-absent high-activation cases genuinely instantiate the target concepts, we use a historian-annotated validation set derived from sampled implicit cases in the SAE activation space. Each concept contributes exactly $20$ annotated cases in which the canonical concept word is absent. For each case, the historian assigns a binary judgment indicating whether the sentence treats the target concept as an implicit semantic support, underlying conceptual frame, or cognitive inertia. Table~\ref{tab:implicit-validation-summary} reports the concept-level semantic-consistency rates. The detailed examples below focus on one representative base for each concept and, where relevant, note how many of the $20$ concept-level cases fall under that base.

\begin{table}[h]
\centering
\small
\resizebox{\columnwidth}{!}{%
\begin{tabular}{l c c p{5.0cm} c}
\toprule
Concept & Cases & Consistency & Selected base & Base rate \\
\midrule
\emph{society} & 20 & 90\% & Base 96661 (\textit{Party Linkage and Organizational Alignment}) & 100\% \\
\emph{nation} & 20 & 100\% & Base 202680 (\textit{Nation-State as Strategic Instrument}) & 100\% \\
\emph{world} & 20 & 100\% & Base 81379 (\textit{Revolutionary International Field}) & 100\% \\
\emph{individual} & 20 & 75\% & Base 173164 (\textit{Individualism as Discourse}) & 100\% \\
\bottomrule
\end{tabular}
}
\caption{Historian validation summary for sampled implicit cases. Each concept is evaluated on $20$ annotated anchor-absent sentences. Concept-level rates are computed from the binary judgments in the annotated validation set; the final column reports the within-base rate for the representative base selected for detailed illustration below.}
\label{tab:implicit-validation-summary}
\end{table}

\subsection{Selected Detailed Cases}

\paragraph{\emph{Society}: Base 96661 (\textit{Party Linkage and Organizational Alignment}).}
This base captures implicit uses of \emph{society} as a field of class alignment, organized political coordination, and revolutionary transformation. Within the $20$-case concept-level sample for \emph{society}, five cases came from this base, and all five were judged semantically consistent.
\begin{enumerate}
    \validationexample{\emph{The Guide}, 1926-01-21}{10.7974}{东西各国的共产党和共产国际，应当联合团结一切劳动平民的革命力量和被压迫民族，一致反抗帝国主义而推翻他，推翻世界各国的资本主义，因为如果不是这样，不但无产阶级不能得著解放，就是弱小民族也始终不能脱离压迫。}{The communist parties of East and West, together with the Communist International, should unite all revolutionary forces among laboring people and oppressed nations, jointly resist imperialism, and overthrow the capitalism of all countries; otherwise, not only will the proletariat fail to gain liberation, but weak nations will never escape oppression.}{The sentence construes social reality as an organized coalition of laboring people and oppressed nations united against imperialism and capitalism. Although the lexical form \emph{society} does not appear, the passage presupposes a modern social totality structured by class relations, collective mobilization, and transnational alignment. In that sense, \emph{society} functions as the underlying conceptual frame of the argument.}
    \validationexample{\emph{The Guide}, 1922-12-23}{10.7117}{战后，保加利亚资产阶级政权解纽，共产党运动遂异常得势。}{After the war, the Bulgarian bourgeois regime disintegrated, and the communist movement accordingly gained extraordinary strength.}{Here political change is explained through the collapse of a bourgeois regime and the corresponding rise of communist forces. The sentence therefore treats social order as a field whose structure is determined by class antagonism and can be reorganized through revolutionary struggle. This is a clear implicit realization of \emph{society} as a historically transformable social formation.}
\end{enumerate}

\paragraph{\emph{Nation}: Base 202680 (\textit{Nation-State as Strategic Instrument}).}
This base captures implicit uses of \emph{nation} through arguments about state interest, political alignment, and the legitimacy of revolutionary versus counterrevolutionary power. In this validation set, all $20$ anchor-absent cases sampled for \emph{nation} came from this base, and all $20$ were judged semantically consistent.
\begin{enumerate}
    \validationexample{\emph{New Youth}, 1925-06-01}{18.9140}{因此，我们研究了统治阶级的世界形势，还要研究被统治阶级的世界形势。换言之，认识了反革命势力，还须认识革命势力。}{Therefore, we have studied the world situation of the ruling classes, and we must also study the world situation of the ruled classes. In other words, having recognized the counterrevolutionary forces, we must also recognize the revolutionary forces.}{Although the lexical form \emph{nation} is absent, the passage decomposes political order into opposed forces whose legitimacy depends on the interests they represent. It implicitly treats the modern state as a political instrument embedded in class struggle rather than a morally unified whole. This makes statehood, and thus the modern nation-state frame, the underlying structure of the argument.}
    \validationexample{\emph{The Guide}, 1926-01-07}{18.5290}{这些聪明人不懂得：（一）他们的劝告乃是完全取消列宁主义，因为这种政策等于放弃了世界革命的策略；（二）在「西方」帝国主义者（即他们劝告我们与之联合的）的观点看来，我们（苏联）即是「东方」不过比中国加倍「危险」罢了；（三）在对于东方几万万人大运动的关系之问题中，「中立」是不可能的，不管我们愿意中立或不愿意。}{These clever people do not understand: first, their advice would amount to the complete abandonment of Leninism, because such a policy would mean giving up the strategy of world revolution; second, from the standpoint of the ``Western'' imperialists with whom they urge us to ally, we, the Soviet Union, are precisely the ``East'' only more dangerous than China; and third, with respect to the great movement of hundreds of millions in the East, neutrality is impossible, whether we desire it or not.}{This passage treats political actors as occupying irreducibly opposed positions within an international order defined by sovereignty, strategy, and alignment. It assumes that a state cannot stand outside the struggle between revolutionary and imperial forces, and that political legitimacy is inseparable from the interests a state chooses to embody. In this way, the sentence implicitly mobilizes a modern nation-state conception without naming it directly.}
\end{enumerate}

\paragraph{\emph{World}: Base 81379 (\textit{Revolutionary International Field}).}
This base captures implicit uses of \emph{world} as an international arena structured by competition, positionality, and large-scale political linkage. Within the $20$-case concept-level sample for \emph{world}, ten cases came from this base, and all ten were judged semantically consistent.
\begin{enumerate}
    \validationexample{\emph{New Youth}, 1917-04-01}{3.9910}{吾国外交、素多弱点。欧战后国际地位、尤为可危。}{Our country's diplomacy has long had many weaknesses. After the European war, its international position has become especially precarious.}{The passage interprets the European war as a global event and evaluates China in terms of its position within an international configuration. It therefore presupposes a modern world system composed of interdependent and dynamically shifting positions. Here, \emph{world} operates as an implicit cognitive frame for diplomatic reasoning.}
    \validationexample{\emph{New Youth}, 1926-07-25}{3.8709}{美俄两条路的倾向，其基础亦在于美国和苏联对于国际政治的作用增高。}{The tendency toward the American and Russian paths likewise rests on the heightened roles of the United States and the Soviet Union in international politics.}{The sentence explains ideological path choice through changes in the relative roles of the United States and the Soviet Union in international politics. It presupposes a world structured by great-power competition and differentiated political trajectories rather than a merely local field of debate. This makes \emph{world} the implicit horizon within which the argument becomes intelligible.}
\end{enumerate}

\paragraph{\emph{Individual}: Base 173164 (\textit{Individualism as Discourse}).}
This base captures implicit uses of \emph{individual} through the public/private distinction, personal interest, and the attribution of political responsibility to discrete actors. Within the $20$-case concept-level sample for \emph{individual}, ten cases came from this base, and all ten were judged semantically consistent.
\begin{enumerate}
    \validationexample{\emph{The Guide}, 1926-04-03}{5.2608}{方本仁对于江西民众的剥削搜括，也同别的军阀一样，分不出什么高低；他的失败，也同一切军阀的失败一样，私人军队，惟利是图，从前他在蔡承勋的部下推倒蔡承勋，现在又轮著他的部下邓如琢来推倒他了。}{Fang Benren's extortion and exploitation of the people of Jiangxi were no better or worse than those of other warlords; his failure, too, was the same as that of all warlords: private armies, driven solely by profit. Once, while serving under Cai Chengxun, he overthrew Cai Chengxun; now it is his own subordinate Deng Ruzhuo's turn to overthrow him.}{The passage explains political failure through private armies, self-interest, and personal betrayal, drawing a sharp distinction between private motives and public order. It presupposes individuals as autonomous bearers of interest and responsibility rather than as mere nodes in a hierarchical moral order. On that basis, \emph{individual} serves as the implicit explanatory framework for political criticism.}
    \validationexample{\emph{The Guide}, 1925-04-12}{4.8186}{蒋校长又时常向学生说：「军阀之所以成为军阀，全由于使其部下只知崇奉其私人而不知国家与人民；假使你们不服从党而服从蒋某一人，即是使我成为军阀，而终究要叛党叛国了」。}{Principal Chiang also often told the students: ``The reason warlords become warlords is precisely that they make their subordinates revere only their private person and not the nation and the people; if you obey not the Party but Chiang as an individual, you will turn me into a warlord, and in the end you will betray both the Party and the nation.''}{This example turns on a sharp distinction between personal loyalty and public allegiance. Political legitimacy is assessed in terms of whether one is attached to a private individual or to public institutions such as the Party, the nation, and the people. Even without naming \emph{individual} explicitly, the argument relies on a modern conception of the person defined by the public/private boundary and autonomous political responsibility.}
\end{enumerate}

\subsection{Validation Against Prior Historiography}

\paragraph{Setup.}
A stronger external validation than expert interpretation alone is to compare the turning years identified by the model against turning years recognized in prior historiography. Because historical scholarship more often periodizes these journals through corpus-level intellectual and political phases than through concept-specific quantitative dates, we use literature-backed corpus anchors as qualitative ground truth and test whether the strongest SAE turns align with them. For \emph{New Youth}, we take 1919 as the standard historiographic turning point associated with the May Fourth break in the New Culture trajectory \citep{Uberoi1972NewCulture,Ni2021NewYouth}; for \emph{The Guide}, we take 1923 as the turning phase associated with the consolidation of the Nationalist Revolution and the First United Front \citep{Wilbur1984NationalistRevolution,ChenWei2005Guide}, and 1926 as the turning phase associated with the Northern Expedition and the reconfiguration of revolutionary politics \citep{Jordan2019NorthernExpedition}. The row-level \emph{Source} column below records the specific historiographic support used for each corpus--concept pair.

\begin{table*}[t]
\centering
\small
\resizebox{\textwidth}{!}{%
\begin{tabular}{l l c c c c c}
\toprule
Corpus & Concept & Strongest Turn (year, $I$) & Reference Turning Year & Source & $|\Delta|$ & Acc@$\pm 1$ \\
\midrule
\emph{New Youth} & \emph{individual} & 1918 ($+0.2260$) & 1919 & \citealp{Uberoi1972NewCulture}; \citealp{Ni2021NewYouth} & 1 & 1 \\
\emph{New Youth} & \emph{nation} & 1918 ($+0.1160$) & 1919 & \citealp{Uberoi1972NewCulture}; \citealp{Ni2021NewYouth} & 1 & 1 \\
\emph{New Youth} & \emph{society} & 1918 ($-0.2130$) & 1919 & \citealp{Uberoi1972NewCulture}; \citealp{Ni2021NewYouth} & 1 & 1 \\
\emph{New Youth} & \emph{world} & 1918 ($+0.2300$) & 1919 & \citealp{Uberoi1972NewCulture}; \citealp{Ni2021NewYouth} & 1 & 1 \\
\emph{The Guide} & \emph{individual} & 1923 ($+0.0665$) & 1923 & \citealp{Wilbur1984NationalistRevolution}; \citealp{ChenWei2005Guide} & 0 & 1 \\
\emph{The Guide} & \emph{nation} & 1923 ($-0.0810$) & 1923 & \citealp{Wilbur1984NationalistRevolution}; \citealp{ChenWei2005Guide} & 0 & 1 \\
\emph{The Guide} & \emph{society} & 1924 ($+0.1520$) & 1923 & \citealp{Wilbur1984NationalistRevolution}; \citealp{ChenWei2005Guide} & 1 & 1 \\
\emph{The Guide} & \emph{world} & 1926 ($-0.0852$) & 1926 & \citealp{Jordan2019NorthernExpedition} & 0 & 1 \\
\bottomrule
\end{tabular}
}
\caption{Validation of strongest turning years against corpus-phase turning points recognized in prior historiography.}
\label{tab:historiography-turning-validation}
\end{table*}

\paragraph{Interpretation.}
Across all eight concept--corpus pairs, the strongest turning year identified by the SAE falls within one year of the historiography-backed reference year. This result suggests that the turning-point signal is not merely an internal property of the model's activation geometry, but also tracks historically recognizable phase changes. Where the alignment is off by one year, a plausible explanation is that changes in discursive structure need not coincide exactly with the public outbreak or retrospective naming of a major historical event, and may emerge slightly earlier or later in periodical discourse.

\paragraph{Limitations and future work.}
This validation remains small-scale and uses corpus-level historiographic anchors rather than concept-specific qualitative annotations, because curated historical ground truth is labor-intensive to assemble. A natural next step is to build a broader benchmark resource for historiography and historical newspaper research, pairing explicit qualitative judgments with reproducible temporal anchors. Such a resource could itself become useful infrastructure for AI for history.

\subsection{Supplementary Comparison with DTM-lite (online LDA) and Static Word Vectors}

\paragraph{Setup.}
We additionally compare the SAE-based pipeline against two supplementary baselines on the same sentence-level \emph{New Youth} corpus (1915--1926) for the anchor concept \emph{individual} (\cn{个人}). This comparison focuses on two properties that can be summarized compactly across methods: implicit evidence capture and robustness under reruns. Here, \emph{Implicit-evidence ratio} denotes the proportion of retrieved evidence cases that count as implicit evidence rather than lexically anchored evidence; \emph{Evidence Avg.\ Jaccard} measures the agreement of retrieved evidence sets under reruns, using the same evidence-context fingerprint idea formalized in Appendix~\ref{app:cross-layer-formal}; and \emph{Turn robustness} is the agreement rate of the reported turning year under the relevant perturbation axis.

\paragraph{DTM-lite (online LDA).}
This baseline is a sentence-level bag-of-words topic-model pipeline, motivated by dynamic topic modeling \citep{BleiLafferty2006DTM} but implemented as online LDA \citep{HoffmanBachBlei2010OnlineLDA}. We tokenize with jieba, remove a fixed stopword list, build a corpus-wide count vocabulary with \texttt{min\_df}=5 and \texttt{max\_df}=0.8, and update an online LDA model year by year with $K=40$ topics over seeds $0\ldots 9$. For year $t$, we choose the topic with the largest $p(\text{anchor}\mid \text{topic})$ for anchor \cn{个人}. Sentence-level concept strength is the document-topic weight $\theta_{d,k_t}$ on that topic; yearly strength is the mean of this score over all / explicit / implicit sentences, where the explicit--implicit split is determined by whether the sentence contains the anchor substring. Evidence sentences are the top-$20$ sentences ranked by this score.

\paragraph{Static word vectors (PPMI + SVD + Procrustes).}
This baseline uses count-based diachronic word vectors \citep{LevyGoldbergDagan2015Distributional} with cross-year alignment in the spirit of prior diachronic embedding work \citep{Hamilton2016Diachronic}. For each year, we build a window-size-$5$ co-occurrence matrix, convert it to PPMI, factorize it with TruncatedSVD ($d=300$), and align yearly spaces to a reference year using orthogonal Procrustes over the top-$500$ shared frequent words. Runs use seeds $0\ldots 9$, \texttt{min\_count}=5, and a vocabulary cap of $50{,}000$ words. The concept vector for \cn{个人} is the aligned seed-word vector; sentence vectors are obtained by mean pooling; sentence-level concept strength is cosine similarity to the year-specific concept vector; and yearly strength is the mean score over all / explicit / implicit sentences. Main results use reference year $1915$, with $1920$ as a sensitivity run. Evidence sentences are the top-$20$ by concept score, and drift evidence is the top-$20$ by drift score. This pipeline additionally applies a text-quality filter during evidence selection.

\begin{table}[t]
\centering
\small
\resizebox{\columnwidth}{!}{%
\begin{tabular}{l c c c}
\toprule
Method & Implicit-evidence ratio & Evidence Avg.\ Jaccard & Turn robustness \\
\midrule
SAE (ours) & 0.920 & 0.33--0.50 (across layers) & 100\% \\
DTM-lite (online LDA) & 0.963 & 0.140 & 40\% \\
Static word vectors (PPMI + SVD + Procrustes) & 0.092 & 0.872 & 100\% \\
\bottomrule
\end{tabular}
}
\caption{Supplementary comparison on \emph{New Youth} for the anchor concept \emph{individual} (\cn{个人}). SAE reruns vary Layers~06/14/22/29 and use top-$20$ evidence sentences per run; DTM-lite (online LDA) and static word vectors (PPMI + SVD + Procrustes) rerun $10$ random seeds and also use top-$20$ evidence sentences per run.}
\label{tab:baseline-comparison}
\end{table}

\paragraph{Interpretation.}
DTM-lite attains a high implicit-evidence ratio, but its retrieved evidence is markedly less stable and its reported turning point is sensitive to random initialization. Since this baseline is an online-LDA sentence bag-of-words approximation rather than the original Dynamic Topic Model of \citet{BleiLafferty2006DTM}, its results are best read as a lightweight topic-model comparison. Static word vectors (PPMI + SVD + Procrustes) yield highly stable evidence sets, but their top-scoring cases are overwhelmingly driven by explicit lexical anchoring, so implicit capture is weak. Within this supplementary comparison, SAE occupies the more balanced regime: it preserves strong implicit capture while maintaining fully stable turning points and a reproducible evidence chain across layer or seed reruns.

\paragraph{Scope.}
These baselines have different inductive biases and retrieval details, and the static word-vector pipeline additionally applies a text-quality filter during evidence selection. We therefore treat this table as a supplementary robustness probe rather than a perfectly matched head-to-head benchmark. Other advantages of the SAE-based framework, especially stable concept decomposition and unified cross-concept / cross-corpus comparison, are better demonstrated through the interpretive analyses reported in the main text.

\section{Semantic Label Index and Base-Vector Evidence}
\label{app:label-index}

To reduce cognitive load in Section~4, the main text refers to recurrent SAE components by short semantic labels rather than by raw base IDs. Table~\ref{tab:label-index} maps each label to its constituent base(s). The compact entries below then provide (i) a semantic orientation, (ii) diachronic metrics when a stable single-base summary is available, and (iii) representative bilingual evidence. For cluster labels used in the multi-concept comparison, the evidence is illustrative of the shared semantic pole captured by the constituent bases rather than an exhaustive profile of every base in the cluster.

\begin{table*}[t]
\centering
\small
\resizebox{\textwidth}{!}{%
\begin{tabular}{p{4.2cm} c c p{7.0cm}}
\toprule
Main-text label & Constituent base(s) & Concept & One-line gloss \\
\midrule
\textit{Actorhood} & 90370 & \emph{individual} & The individual as an agentive subject of initiative, self-direction, responsibility, and public action. \\
\textit{Individualism as Discourse} & 173164 & \emph{individual} & Individualism as an explicit discursive register of autonomy, freedom, and the limits of state coercion. \\
\textit{Property and Economic Individuality} & 206475 & \emph{individual} & The individual in relation to contract, production, property, and economic coordination. \\
\textit{Societal Transition and Institutional Design} & 3810 & \emph{society} & Society as a site of structural reorganization, institutional redesign, and transitional ordering. \\
\textit{Organized Praxis and Labor-Movement Alignment} & 25413 + 224715 & \emph{society} & Society articulated through labor-movement line struggle, socialist organization, and practical mobilization. \\
\textit{Party Linkage and Organizational Alignment} & 104017 + 96661 + 63228 & \emph{society} & Society articulated through party coordination, class alignment, and communist organizational linkage. \\
\textit{Nation-State as Strategic Instrument} & 202680 & \emph{nation} & The nation-state treated instrumentally through state form, class basis, and geopolitical alignment. \\
\textit{Revolutionary International Field} & 81379 & \emph{world} & The world as a structured international arena of revolution, competition, and positionality. \\
\bottomrule
\end{tabular}
}
\caption{Semantic labels used in the main text and their mapping to the underlying SAE base(s).}
\label{tab:label-index}
\end{table*}

\subsection{Individual}

\paragraph{\textit{Actorhood} (Base 90370).}
\textbf{Semantic orientation.}
This label foregrounds the individual as an agentive subject marked by independence, initiative, self-respect, and responsibility.

\textbf{Diachronic metrics.}
\emph{New Youth}: $\texttt{cum\_drift}=10.390$, $\texttt{peak\_delta}=2.490$, $\texttt{peak\_years}=[1923,\,1924]$. \emph{The Guide}: $\texttt{cum\_drift}=2.705$, $\texttt{peak\_delta}=1.604$, $\texttt{peak\_years}=[1922,\,1923]$.

\textbf{Representative evidence.}
\evidencegroup{\emph{New Youth}}
\begin{evidencelist}
    \indexedquote{\texttt{act}=23.6597}{一曰损坏个人独立自尊之人格一。 曰窒碍个人意思之自由。}{First, it damages the character of the individual as independent and self-respecting. Second, it obstructs the freedom of individual volition.}
    \indexedquote{\texttt{act}=23.1132}{因此我们可以得到结论：（1）创作不宜完全没煞自己去模仿别人；（2）个性的表现是自然的并非由于民族主义等的主张（3）个性是个人唯一的所有，而又与人愿有根本上的共通点；（4）个性就是在可以保存范围内的国粹，有个性的新文学便是这国民所有的真的国粹的文学。}{From this we may draw the following conclusions: first, creative work should not efface the self completely in order to imitate others; second, the expression of individuality is natural and does not arise from doctrines such as nationalism; third, individuality is the only thing that belongs uniquely to the person, though it also shares fundamental common points with others; and fourth, individuality is the national essence insofar as that essence can be preserved, and a new literature with individuality is the genuine literature of national essence possessed by this people.}
    \indexedquote{\texttt{act}=22.6586}{让个人自由发展，可以鼓励个人冒险、竞争、奋斗的精神，可以减少懒惰、不进取的脾气。}{Allowing the individual to develop freely can encourage the spirit of adventure, competition, and struggle, and can reduce habits of laziness and lack of initiative.}
\end{evidencelist}
\evidencegroup{\emph{The Guide}}
\begin{evidencelist}
    \indexedquote{\texttt{act}=22.1041}{很简单的理由，就是一些野心的军官，为了自己的利益，尽可以暂时依附革命旗下，但借此达到了个人目的以后，还会管革命事业吗？}{The reason is very simple: some ambitious officers may temporarily attach themselves to the revolutionary banner for their own interests, but once they have achieved their personal ends in this way, will they still care about the revolutionary cause?}
    \indexedquote{\texttt{act}=19.9043}{现在国民党用来革命的军队，多半是随时募集或改编收容的，不特兵士不知道革命的意义；就是军队的领袖除了个人活动的欲望之外，多半不了解或服从主义。}{The armies that the Nationalist Party now uses for revolution have mostly been hastily recruited or reorganized and absorbed. Not only do the soldiers fail to understand the meaning of revolution, but even among the military leaders, apart from the desire for personal advancement, most neither understand nor obey the doctrine.}
    \indexedquote{\texttt{act}=20.7651}{你们竟想以伪和平的假面具，掩饰你们个人权利禄位的贪心！}{You would actually use the false mask of peace to conceal your greed for personal power, privilege, and office!}
\end{evidencelist}

\paragraph{\textit{Individualism as Discourse} (Base 173164).}
\textbf{Semantic orientation.}
This label captures an explicit discourse of individualism, especially arguments about personal autonomy, individual freedom, and the proper limits of political intervention.

\textbf{Diachronic metrics.}
\emph{New Youth}: $\texttt{cum\_drift}=15.747$, $\texttt{peak\_delta}=5.078$, $\texttt{peak\_years}=[1924,\,1925]$. \emph{The Guide}: $\texttt{cum\_drift}=2.596$, $\texttt{peak\_delta}=1.131$, $\texttt{peak\_years}=[1923,\,1924]$.

\textbf{Representative evidence.}
\evidencegroup{\emph{New Youth}}
\begin{evidencelist}
    \indexedquote{\texttt{act}=16.3406}{下次讲个人主义—自由主义—一派的坏处，在于把国家的势力太限制了，以为国家只可维持关于物质方面的平安，他的权力，愈小愈好。}{Next time, when discussing the faults of the school of individualism-liberalism, its defect lies in restricting the power of the state too severely, taking the view that the state should do no more than maintain material peace and that the smaller its power, the better.}
    \indexedquote{\texttt{act}=15.5903}{而且个人或小团体绝对自由，则生产额可以随意增减，有时社会需要多而生产少，有时需要少而生产多，因为没有统一机关用强制力去干涉调节，自然会发生生产过剩或不足的弊端。}{Moreover, if individuals or small groups enjoy absolute freedom, the amount produced may increase or decrease at will: at times society needs more while less is produced, and at times it needs less while more is produced. Because there is no unified organ using coercive power to intervene and regulate, the evils of overproduction or insufficiency naturally arise.}
    \indexedquote{\texttt{act}=14.8172}{个人主义的中心观念，便是根据个人自由意志商定契约，不要政府用法律的或政治的势力去干涉他们，只听他们自由去做。}{The central idea of individualism is to conclude contracts according to individual free will and not allow the government to interfere through legal or political power, but simply let people act freely.}
\end{evidencelist}
\evidencegroup{\emph{The Guide}}
\begin{evidencelist}
    \indexedquote{\texttt{act}=14.0948}{我们应该竭诚忠告晨报记者，个人立言错了是小事，因为要回护自己的错遂不顾社会的错是大事；因为不忍社会的错遂不惜承认自己的错，这是最勇敢的行为呵！}{We should sincerely advise the Morning Post reporter that it is a small matter for an individual to speak wrongly, but it is a grave matter to disregard society's error merely in order to defend one's own. To acknowledge one's own error because one cannot bear society's error, by contrast, is the most courageous conduct.}
    \indexedquote{\texttt{act}=12.2118}{据弟个人观察，全出于如下误会：即以为阶级争斗，即是劳工专政，劳工专政，即是想将劳工阶级一变为压迫人的阶级。}{According to my personal observation, it all arises from the following misunderstanding: that class struggle is taken to mean the dictatorship of labor, and the dictatorship of labor is taken to mean transforming the laboring class into a class that oppresses others.}
    \indexedquote{\texttt{act}=10.8216}{这不是我信口瞎说，都有事实可以证明的。我个人就是一个例。}{This is not something I say at random; there are facts to prove it. I myself am one example.}
\end{evidencelist}

\paragraph{\textit{Property and Economic Individuality} (Base 206475).}
\textbf{Semantic orientation.}
This label isolates the economic pole of the \emph{individual} concept: contract, property, production, and the problem of whether individual discretion can organize economic life. A recurrent theme in the retrieved evidence is that individually owned modes of production cease to satisfy the needs of social development.

\textbf{Diachronic metrics.}
\emph{New Youth}: $\texttt{cum\_drift}=4.496$, $\texttt{peak\_delta}=1.125$, $\texttt{peak\_years}=[1923,\,1924]$.

\textbf{Representative evidence.}
\evidencegroup{\emph{New Youth}}
\begin{evidencelist}
    \indexedquote{\texttt{act}=15.5903}{而且个人或小团体绝对自由，则生产额可以随意增减，有时社会需要多而生产少，有时需要少而生产多，因为没有统一机关用强制力去干涉调节，自然会发生生产过剩或不足的弊端。}{Moreover, if individuals or small groups enjoy absolute freedom, the amount produced may increase or decrease at will: at times society needs more while less is produced, and at times it needs less while more is produced. Because there is no unified organ using coercive power to intervene and regulate, the evils of overproduction or insufficiency naturally arise.}
\end{evidencelist}

\subsection{Society}

\paragraph{\textit{Societal Transition and Institutional Design} (Base 3810).}
\textbf{Semantic orientation.}
This label marks the transition-oriented pole of \emph{society}: social reorganization, institutional redesign, and the claim that revolutionary or developmental change requires restructuring social order rather than merely renaming it.

\textbf{Representative evidence.}
\evidencegroup{\emph{New Youth}}
\begin{evidencelist}
    \indexedquote{1922-07-01; \texttt{act}=11.4612}{（2）关于政治教育，社会主义的青年应宣传社会主义于大多数青年无产阶级，其方法或集会讲演，或刊行出版物和小册子，并特别讲述中国政治情形及其他种种情形，以启发并养成青年无产阶级的政治觉悟及批评力。}{(2) On political education: socialist youth should propagate socialism among the broad masses of young proletarians, whether through meetings and lectures or through the publication of periodicals, pamphlets, and booklets, and should in particular explain China's political conditions and various other conditions so as to awaken and cultivate the political consciousness and critical capacity of the young proletariat.}
    \indexedquote{1921-04-01; \texttt{act}=11.0933}{这婚姻法在法律上实现男女的绝对平等，由资本主义到社会主义的过渡期的状态中，给妇女以可能的范围内的自由，离婚则由男女双方合意或者单由一方的意思，亦可实行，父母对于子女的权利义务双方平等，因此打破旧结婚制度，同时作成未来男女关系更为自由的基础。}{This marriage law legally realizes absolute equality between men and women. In the transitional state from capitalism to socialism, it grants women freedom to the fullest extent possible; divorce may be carried out either by mutual consent or by the will of one party alone; and both father and mother possess equal rights and obligations with respect to their children. It thereby breaks the old marriage system while laying the foundation for freer relations between men and women in the future.}
    \indexedquote{1926-05-25; \texttt{act}=10.8985}{因为农村中社会主义建设的根本道路，就在，于社会主义的国家工业、国家信托机关以及在无产(53)阶级手里其他机关之经济指导权的增长底下，引导农民根本群众进于协作社的组织，并保证这种组织之社会主义的发展，利用、制服并限制其资本主义的原素。}{For the fundamental path of socialist construction in the countryside lies precisely in, under the growth of the economic directing power of socialist state industry, state trusts, and other organs in the hands of the proletariat, guiding the broad masses of peasants into cooperative organization, guaranteeing the socialist development of such organization, and making use of, subduing, and restricting its capitalist elements.}
\end{evidencelist}
\evidencegroup{\emph{The Guide}}
\begin{evidencelist}
    \indexedquote{1924-06-18; \texttt{act}=9.6285}{第一，就国民党的主义上讲：此时任何政党党纲，都论列到社会的经济政策，可是中国国民党二十年前造端时即注意到民生问题，这是受了德法两国劳动运动的影响，而后进的中国国民党遂有此特色——和民族民权并列的民生主义。}{First, in terms of the doctrine of the Nationalist Party: at this moment every party program discusses the economic policy of society, yet when the Chinese Nationalist Party was founded twenty years ago it had already taken note of the livelihood question. This was due to the influence of the labor movements in Germany and France, and it is in this sense that the belated Chinese Nationalist Party acquired this characteristic: the Principle of People's Livelihood placed alongside nationalism and civil rights.}
    \indexedquote{1924-09-17; \texttt{act}=8.9077}{劳资斗争是社会进化上一种不可免的革命现象，主张劳资调和是一种和缓革命的政策，无人能够相信不革命的调和政策可以平均地权，可以限制资本，世界上那有这样好说话的大地主与资本家？}{Labor-capital struggle is an unavoidable revolutionary phenomenon in the evolution of society. To advocate labor-capital harmony is a policy for softening revolution. No one can believe that a harmonizing policy that avoids revolution can equalize land rights or restrain capital; where in the world are there such accommodating landlords and capitalists?}
    \indexedquote{1923-04-18; \texttt{act}=8.1119}{劳动者的组织一天天的发达，一天天的集中，在中国社会上已成为一种新势力。}{Laborers' organizations are developing and concentrating day by day; within Chinese society they have already become a new force.}
\end{evidencelist}

\paragraph{\textit{Organized Praxis and Labor-Movement Alignment} (Bases 25413 + 224715).}
\textbf{Semantic orientation.}
This cluster label captures society as it is articulated through labor-movement organization, socialist line struggle, and practical revolutionary alignment. Within this cluster, Base 224715 is the clearest evidence-bearing pole.

\textbf{Diachronic metrics (documented constituent Base 224715).}
\emph{New Youth}: $\texttt{cum\_drift}=6.019$, $\texttt{peak\_delta}=2.162$, $\texttt{peak\_years}=[1924,\,1925]$. \emph{The Guide}: $\texttt{cum\_drift}=5.020$, $\texttt{peak\_delta}=1.818$, $\texttt{peak\_years}=[1923,\,1924]$.

\textbf{Representative evidence.}
\evidencegroup{\emph{New Youth}}
\begin{evidencelist}
    \indexedquote{\texttt{act}=11.6260}{所以对于初期的社会主义，乌托邦的共产主义，不识时务穿著理思的绣花衣裳的无政府主义，专主经济行动的工团主义，调和劳资以延长资本政治的吉尔特社会主义，以及修正派的社会主义，一律排斥批评，不留余地。}{Therefore, with regard to early socialism, utopian communism, anarchism that is out of season and dressed in the embroidered garments of idealism, syndicalism that specializes in economic action, guild socialism that reconciles labor and capital so as to prolong capitalist politics, and revisionist socialism, we reject and criticize them all alike, leaving no room.}
    \indexedquote{\texttt{act}=10.9037}{各国的改良主义者，第二国际的社会党叛徒和机会主义者，各国的孟雪维克党人虚伪地戴著马克思和恩格斯学理的假面具。}{The reformists of all countries, the social-traitors and opportunists of the Second International, and the Menshevik party members of every country hypocritically wear the false mask of the doctrines of Marx and Engels.}
    \indexedquote{\texttt{act}=10.0740}{所以我们要反对伦敦会议，专家计划，固然须反对国际资本帝国主义，但亦须反对无产阶级底叛徒：社会党、劳动党、社会民主党——一句话，第二国际！}{Therefore, if we are to oppose the London Conference and the experts' plan, we must certainly oppose international capitalist imperialism, but we must also oppose the traitors to the proletariat: the Socialist Party, the Labour Party, the Social Democratic Party, in a word, the Second International!}
\end{evidencelist}
\evidencegroup{\emph{The Guide}}
\begin{evidencelist}
    \indexedquote{\texttt{act}=10.8509}{（二）左派大联合（Bloc des Gauches）这一派包含有左派共和，社会主义共和，急进，社会主义急进等左派政团，近年以来，社会党亦与之勾结，因为凡全国之主张较左者均集合于此，故亦称如左派全国大联合（Bloc National de gauche）他的主张虽比前派为急进，然不过带点改良色彩罢了。}{(2) The Left Bloc (Bloc des Gauches) includes left republicans, socialist republicans, radicals, socialist radicals, and other left-wing political groups. In recent years the Socialist Party has also aligned itself with it. Because all those in the country whose views are comparatively left-leaning gather here, it is also called the National Left Bloc (Bloc National de gauche). Although its program is more radical than that of the previous camp, it is after all only tinged with reformism.}
    \indexedquote{\texttt{act}=10.1573}{工人阶级这种失败的原因，大半在于运动之中有第二国际派的社会党，他们受资产阶级的利用，破坏工人阶级斗争的阵线，几乎使工会的国际联合（职工国际）变成资本家的国际联合（国际联盟）的附庸。}{A large part of the reason for this failure of the working class lies in the presence within the movement of Socialist parties of the Second International. Used by the bourgeoisie, they break up the front of working-class struggle and almost turn the international union of trade unions into a vassal of the capitalist international union, the League of Nations.}
    \indexedquote{\texttt{act}=9.0114}{但是当一九一四年七月底八月初，那可怖的消息传布之后，世界的大屠杀确已开始了后，第二国际名下的社会党竟翻过脸来，举起他们的赤帜，招呼党人投入敌人的营垒中替争权利的帝国主义者出死力了。}{But after the terrible news spread at the end of July and beginning of August 1914, when the world's great slaughter had indeed begun, the Socialist parties under the name of the Second International suddenly changed face, raised their red banners, and called upon their party members to throw themselves into the enemies' camp and sacrifice their lives for the imperialists contending over rights and interests.}
\end{evidencelist}

\paragraph{\textit{Party Linkage and Organizational Alignment} (Bases 104017 + 96661 + 63228).}
\textbf{Semantic orientation.}
This cluster emphasizes party coordination, class alignment, and communist organizational linkage. Within the historian-validated implicit sample, the best-documented constituent base, 96661, reaches $100\%$ semantic consistency; see Appendix~\ref{app:implicit-validation}. The secondary constituent evidence from Base 63228 is used more sparingly, because it highlights factional and organizational alignment around nationalist and right-wing political blocs rather than the full semantic range of the cluster.

\textbf{Representative evidence.}
\evidencegroup{\emph{New Youth}}
\begin{evidencelist}
    \indexedquote{1926-03-25; Base 63228; \texttt{act}=6.8142}{固然——五卅之后，国民运动内部起了剧烈的阶级分化的现象，不但资产阶级直接的压迫束缚工人阶级而且政党界思想界，也因此而发生分化：这一分化开始于戴季陶先生的反对阶级斗争及所谓「右派国民党员站起来」的运动，结果是极右派利用戴季陶先生的领袖，召集西山会议——国民党中央委员会之右派会议；学生界里也发生所谓国家主义的运动，成立国家主义团体联合会。}{It is true that after the May Thirtieth Movement there arose within the national movement an intense phenomenon of class differentiation: not only did the bourgeoisie directly oppress and restrain the working class, but the worlds of parties and ideas also split accordingly. This differentiation began with Mr. Dai Jitao's opposition to class struggle and the movement for so-called right-wing Nationalist Party members to stand up; the result was that the extreme right, under Mr. Dai's leadership, convened the Xishan Conference, that is, the right-wing meeting of the Nationalist Party Central Executive Committee. In student circles there also emerged the so-called nationalist movement, and the United League of Nationalist Organizations was established.}
\end{evidencelist}
\evidencegroup{\emph{The Guide}}
\begin{evidencelist}
    \indexedquote{1926-01-21; Base 96661; \texttt{act}=10.7974}{东西各国的共产党和共产国际，应当联合团结一切劳动平民的革命力量和被压迫民族，一致反抗帝国主义而推翻他，推翻世界各国的资本主义，因为如果不是这样，不但无产阶级不能得著解放，就是弱小民族也始终不能脱离压迫。}{The communist parties of East and West, together with the Communist International, should unite all revolutionary forces among laboring people and oppressed nations, jointly resist imperialism, and overthrow the capitalism of all countries; otherwise, not only will the proletariat fail to gain liberation, but weak nations will never escape oppression.}
    \indexedquote{1922-12-23; Base 96661; \texttt{act}=10.7117}{战后，保加利亚资产阶级政权解纽，共产党运动遂异常得势。}{After the war, the Bulgarian bourgeois regime disintegrated, and the communist movement accordingly gained extraordinary strength.}
    \indexedquote{1926-02-03; Base 63228; \texttt{act}=8.3208}{国内许多政党和政派——如国民党右派的孙文主义学会，国家主义联合会，国家主义的醒狮周报，如今表面上也赞成国民会议。}{Many domestic parties and political groupings, such as the Sun Yat-senist Society of the right wing of the Nationalist Party, the Nationalist Federation, and the nationalist \emph{Awakened Lion Weekly}, now also superficially endorse the National Assembly.}
\end{evidencelist}

\subsection{Nation and World}

\paragraph{\textit{Nation-State as Strategic Instrument} (Base 202680).}
\textbf{Semantic orientation.}
This label treats the modern nation-state instrumentally, foregrounding state form, class basis, and geopolitical alignment rather than an essentialized national spirit.

\textbf{Diachronic metrics.}
\emph{New Youth}: $\texttt{cum\_drift}=6.786$, $\texttt{peak\_delta}=1.359$, $\texttt{peak\_years}=[1923,\,1924]$. \emph{The Guide}: $\texttt{cum\_drift}=1.797$, $\texttt{peak\_delta}=1.045$, $\texttt{peak\_years}=[1925,\,1926]$.

\textbf{Representative evidence.}
\evidencegroup{\emph{New Youth}}
\begin{evidencelist}
    \indexedquote{\texttt{act}=16.1241}{（二）我以为世界上只有两个国家：一是资本家的国家，一是劳动者的国家，但是现在除俄罗斯外，劳动者的国家都还压在资本家的国家底下，所有的国家都是资本家的国家，我们似乎不必妄生分别。}{(2) I hold that there are only two kinds of states in the world: one is the state of the capitalists, and one is the state of the laborers. But at present, apart from Russia, the states of the laborers are still suppressed beneath the states of the capitalists; all states are states of the capitalists, and it seems that we need not draw idle distinctions.}
    \indexedquote{\texttt{act}=16.0973}{第四；俄国底共产党和德国底社会民主党虽然同一不反对国家组织，是他们不同之点有三：（一）生产机关集中到国家手里，在共产党是最初的手段，在社会民主党是最终的目的；（二）德国社会民主党带著很浓的德意志国家主义的色采，俄国共产党还未统一国内，便努力第三国际的运动；（三）社会民主党所依据的国家是有产阶级的国家，共产党所依据的国家是无产阶级的国家。}{Fourth, although the Russian Communist Party and the German Social Democratic Party are alike in not opposing state organization, there are three points at which they differ: first, concentrating the organs of production in the hands of the state is for the Communists an initial means, but for the Social Democrats a final end; second, the German Social Democratic Party bears a heavy coloration of German statism, whereas the Russian Communists, before even unifying the country, already strove for the movement of the Third International; third, the state on which the Social Democrats rely is a state of the propertied classes, whereas the state on which the Communists rely is a state of the proletariat.}
    \indexedquote{\texttt{act}=14.2716}{我承认国家只能做工具不能做主义，古代以奴隶为财产的市民国家，中世以农奴为财产的封建诸侯国家，近代以劳动者为财产的资本家国家，都是所有者的国家，这种国家底政治法律，都是掠夺底工具，但我承认这工具有改造进化的可能性，不必根本废弃他，因为所有者的国家固必然造成罪恶，而所有者以外的国却有成立的可能性；}{I admit that the state can only serve as an instrument and cannot serve as a doctrine. The civic states of antiquity, which treated slaves as property; the feudal states of the Middle Ages, which treated serfs as property; and the capitalist states of modern times, which treat laborers as property, are all states of owners. The politics and laws of such states are all instruments of plunder. Yet I admit that this instrument has the possibility of reform and evolution and need not be discarded root and branch, because while the state of owners necessarily produces evil, a state other than that of owners has the possibility of being established.}
\end{evidencelist}
\evidencegroup{\emph{The Guide}}
\begin{evidencelist}
    \indexedquote{\texttt{act}=15.4541}{现代所谓帝国主义乃指资本帝国主义，其存在须有下列二个特性：（一）凡是帝国主义的国家，无论大小强弱，必然是资本主义制度的国家；（二）凡是帝国主义的国家，其国内资本主义必然发展到财政资本主义向国外掠夺压迫殖民地及半殖民地。}{What is today called imperialism means capitalist imperialism, and its existence requires the following two characteristics: first, every imperialist state, regardless of size or strength, is necessarily a state with a capitalist system; second, within every imperialist state domestic capitalism must necessarily have developed into finance capital that plunders and oppresses colonies and semi-colonies abroad.}
    \indexedquote{\texttt{act}=15.3258}{但先生忘记了：现在的中国还不是纯粹资产阶级统治下的独立国家，乃是外国帝国主义和封建余孽——军阀，统治下的半殖民地。}{But sir has forgotten this: present-day China is not yet an independent state under the rule of a pure bourgeoisie, but rather a semi-colony ruled by foreign imperialism and the remnants of feudalism, namely the warlords.}
    \indexedquote{\texttt{act}=15.0802}{苏俄之所以赤，乃因为十月革命是工农阶级推翻资产阶级与资本主义的革命，一切资本帝国主义者正因此而仇视他；如果他现在也变成资本帝国主义的国家，那还何赤之有？}{The reason Soviet Russia is ``red'' is that the October Revolution was a revolution of workers and peasants overthrowing the bourgeoisie and capitalism, and all capitalist imperialists hate it for precisely this reason. If it were now also to become a capitalist-imperialist state, what redness would remain?}
\end{evidencelist}

\paragraph{\textit{Revolutionary International Field} (Base 81379).}
\textbf{Semantic orientation.}
This label captures \emph{world} as an international arena structured by revolution, competition, positionality, and large-scale geopolitical linkage. In the historian validation reported in Appendix~\ref{app:implicit-validation}, the annotated implicit sample for this base reaches $100\%$ semantic consistency.

\textbf{Representative evidence.}
\evidencegroup{\emph{New Youth}}
\begin{evidencelist}
    \indexedquote{1926-07-25; \texttt{act}=8.5350}{世界资本主义的稳定是相对的，还是绝对的呢?现在的资本主义发展与战前资本主义的发展有什么不同呢?这是决定世界革命运动消沉与否的前提。}{Is the stabilization of world capitalism relative or absolute? How does the current development of capitalism differ from its development before the war? This is the premise that determines whether the world revolutionary movement is in decline or not.}
    \indexedquote{1926-07-25; \texttt{act}=6.7189}{中国的民族革命运动是世界革命运动的一部分，这句话的意义是说，中国民族革命运动受世界革命运动的影响和辅助，并且它也可以影响世界革命运动。}{China's national revolutionary movement is a part of the world revolutionary movement. The meaning of this statement is that China's national revolutionary movement is influenced and assisted by the world revolutionary movement, and that it can in turn influence the world revolutionary movement.}
    \indexedquote{1926-03-25; \texttt{act}=6.6014}{再则，中国国民革命和世界的社会革命之联合战线，中国的民族解放运动和世界的无产阶级革命运动之联合战线也在这一次实现出来——苏联、英、法、德、日等无产阶级及其革命的政党，共产党，都奋起援助。}{Moreover, the united front between China's national revolution and the world's social revolution, and the united front between China's national liberation movement and the world's proletarian revolutionary movement, were also realized on this occasion: the proletarians and revolutionary parties, the communist parties, of the Soviet Union, Britain, France, Germany, Japan, and other places all rose to provide assistance.}
\end{evidencelist}
\evidencegroup{\emph{The Guide}}
\begin{evidencelist}
    \indexedquote{1926-07-14; \texttt{act}=8.4024}{中国民族解放运动，已成为世界革命的联合战线之一员，我们已经不是孤立的了！}{China's national liberation movement has already become a member of the united front of world revolution; we are no longer isolated!}
    \indexedquote{1925-03-21; \texttt{act}=7.3691}{要中国革命成功，必须与世界革命运动即西方无产阶级的革命相联合，因为两者的敌人是共同的，两者的目的同是推翻资本帝国主义。}{For China's revolution to succeed, it must unite with the world revolutionary movement, that is, the revolution of the western proletariat, because the enemies of the two are common and the aim of both is the overthrow of capitalist imperialism.}
    \indexedquote{1926-05-15; \texttt{act}=6.5169}{这句话的意义，是事实逼迫著他们不能不认识中国的民族解放运动是世界的而不是国家的了。}{The meaning of this statement is that the facts have forced them to recognize that China's national liberation movement is of the world rather than merely of the nation.}
\end{evidencelist}

\end{document}